\documentclass[runningheads]{llncs}
\usepackage[T1]{fontenc}
\usepackage{graphicx}
\usepackage{booktabs}
\usepackage[misc]{ifsym}

\usepackage{mwe}

\usepackage{dsfont}
\usepackage{booktabs}
\usepackage{multirow}
\usepackage{makecell}

\usepackage{newfloat}
\usepackage{listings}

\usepackage[dvipsnames]{xcolor}

\usepackage[utf8]{inputenc}
\usepackage{hyperref}
\usepackage{multirow}
\usepackage{url}
\usepackage{physics}
\usepackage{xcolor}
\usepackage{amsmath, amsfonts}
\usepackage{graphicx}
\usepackage{thmtools,thm-restate}
\usepackage{subcaption}
\usepackage[capitalize,noabbrev]{cleveref}
\usepackage{algorithmic}
\usepackage{algorithm}
\usepackage[misc]{ifsym}

\DeclareMathOperator*{\argmin}{argmin}

\spnewtheorem{assumption}{Assumption}{\bfseries}{\itshape}
\spnewtheorem{prop}{Proposition}{\bfseries}{\itshape}
\spnewtheorem{sublemma}{Lemma}[subsection]{\bfseries}{\itshape}
\spnewtheorem{subcorollary}[sublemma]{Corollary}{\bfseries}{\itshape}

\definecolor{codegreen}{rgb}{0,0.6,0}
\definecolor{codegray}{rgb}{0.5,0.5,0.5}
\definecolor{codepurple}{rgb}{0.58,0,0.82}
\definecolor{backcolour}{rgb}{0.95,0.95,0.92}

\lstdefinestyle{mystyle}{
    backgroundcolor=\color{backcolour},   
    commentstyle=\color{codegreen},
    keywordstyle=\color{magenta},
    numberstyle=\tiny\color{codegray},
    stringstyle=\color{codepurple},
    basicstyle=\ttfamily\footnotesize,
    breakatwhitespace=false,         
    breaklines=true,                 
    captionpos=b,                    
    keepspaces=false,                 
    numbers=left,                    
    numbersep=2pt,                  
    showspaces=false,                
    showstringspaces=false,
    showtabs=false,                  
    tabsize=2
}
\begin{document}

\title{Hybrid Cross-domain Robust Reinforcement Learning}


\author{Linh Le Pham Van\inst{1} \and Minh Hoang Nguyen\inst{1} \and Hung Le\inst{1} \and Hung The Tran\inst{2} \and Sunil Gupta\inst{1}}

\authorrunning{L. Le Pham Van et al.}

\institute{Deakin Applied Artificial Intelligence Initiative, Deakin University, Australia \email{\{l.le, s223669184, thai.le, sunil.gupta\}@deakin.edu.au}
\and
Hanoi University of Science and Technology, Hanoi, Vietnam \email{hungtt@soict.hust.edu.vn}}

\maketitle              

\begin{abstract}
Robust reinforcement learning (RL) aims to learn policies that remain effective despite uncertainties in its environment, which frequently arise in real-world applications due to variations in environment dynamics. The robust RL methods learn a robust policy by maximizing value under the worst-case models within a predefined uncertainty set. Offline robust RL algorithms are particularly promising in scenarios where only a fixed dataset is available and new data cannot be collected. However, these approaches often require extensive offline data, and gathering such datasets for specific tasks in specific environments can be both costly and time-consuming. Using an imperfect simulator offers a faster, cheaper, and safer way to collect data for training, but it can suffer from dynamics mismatch. In this paper, we introduce HYDRO, the first Hybrid Cross-Domain Robust RL framework designed to address these challenges. HYDRO utilizes an online simulator to complement the limited amount of offline datasets in the non-trivial context of robust RL. By measuring and minimizing performance gaps between the simulator and the worst-case models in the uncertainty set, HYDRO employs novel uncertainty filtering and prioritized sampling to select the most relevant and reliable simulator samples. Our extensive experiments demonstrate HYDRO's superior performance over existing methods across various tasks, underscoring its potential to improve sample efficiency in offline robust RL.

\end{abstract}

\section{Introduction}

Reinforcement learning (RL) has shown remarkable success in real-world applications \cite{mnih2015human,schrittwieser2020mastering}, but deploying RL policies is often challenged by fluctuations in environment dynamics. Many existing methods assume consistency between training and deployment environments, an assumption frequently violated in practice due to such fluctuations. For instance, a robot operating in a dynamic real-world environment may encounter variations in mass, friction, and sensor noise compared to its training environment, leading to performance degradation \cite{peng2018sim,sunderhauf2018limits}. The Robust Markov Decision Process (RMDP) framework \cite{iyengar2005robust} addresses this challenge by modeling uncertain test environments as a set of possible models around a training model, which is often called the nominal model.
Robust RL aims to learn an optimal policy that maximizes performance under the worst-case scenario within this uncertainty set, using only the nominal model.

\begin{figure} 
    \centering
    \includegraphics[width=0.5\linewidth]{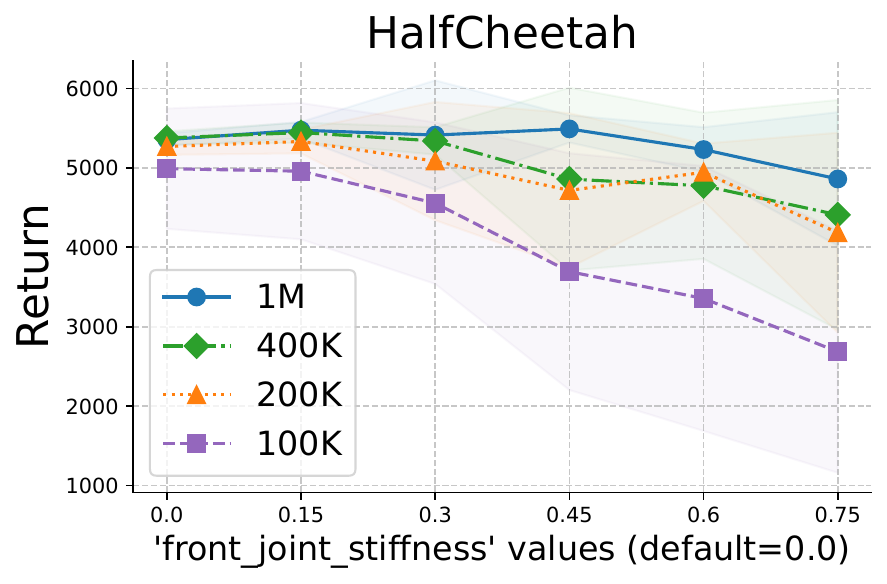}
    \caption{Problem of existing offline robust RL model: Robustness performance drops significantly when training data decreases. Figure illustrates performance comparison under `front\_joint\_stiffness' perturbation of offline robust RL \cite{panaganti2022robust} with different training data sizes from HalfCheetah medium dataset (D4RL).}
    \label{fig:teaser}
\end{figure}

Since its introduction \cite{iyengar2005robust,nilim2005robust}, the RMDP framework has been extensively studied in the context of planning problems \cite{xu2010distributionally}. 
Recently, many robust RL algorithms, learning robust policies from unknown nominal models, have also been proposed \cite{wang2021online,dong2022online}. Still, all these works are limited to the online setting, where policy learning requires online interactions with the environment. Recent success in offline RL \cite{levine2020offline,kumar2020conservative,lyu2022mildly} has motivated the development of offline robust RL methods  \cite{panaganti2022robust,shi2022distributionally,ma2022distributionally,blanchet2024double} to alleviate this restriction.
Despite this progress, offline robust RL methods rely on large datasets, and current offline robust RL struggles when the amount of training data is reduced.
As shown in Figure \ref{fig:teaser}, the robustness performance of the robust RL method drops significantly when the amount of data decreases. This raises a natural question: \emph{Can we reduce the required offline training data without sacrificing the performance of the learn policy under uncertain deployment environments?} 

To address the challenge of scarce data in offline RMDP, we propose a method that utilizes a simulator (source domain), an imperfect model but a faster, cheaper, and safer place to collect data for training the agent. We aim to leverage the interaction with this simulator to mitigate performance degradation caused by the limited offline nominal (target) dataset. A simulator enables unrestricted exploration and access to abundant, diverse data, potentially compensating for limited coverage of offline datasets.  
The potential of using an additional source environment to bring sample efficiency has sparked research interest with numerous methods proposed in the Markov Decision Process (MDP) setting, often referred to as cross-domain RL \cite{niu2024comprehensive}. However, naively combining source and target data may lead to performance degradation due to dynamics mismatches \cite{wencontrastive}. Thus, in the normal MDP setting, prior methods measure the domain gap directly between target and source domains using corresponding datasets. 
This approach cannot be directly applied to RMDPs where we optimize for \emph{worst-case} performance while only having access to nominal model data. To the best of our knowledge, our work is the first to study the combination of online source and offline nominal dataset in the robust RL setting. 

In this paper, we introduce HYDRO, the first Hybrid Cross-Domain Robust RL framework. HYDRO measures the performance gap caused by the dynamics mismatch between the simulator and the worst-case model in an uncertainty set around the nominal target model. Using this measurement, HYDRO uses a novel uncertainty based filtering mechanism and priority sampling scheme to select reliable and relevant samples from the simulator, minimizing performance degradations. Our contributions are:
\begin{itemize}
    \item We are the first to address the Hybrid cross-domain Robust RL setting, which is a novel problem, and develop a method to improve the sample efficiency for the offline robust RL. 
    \item We perform a theoretical analysis of our novel problem setting and use it to propose HYDRO, a practical and effective algorithm for solving this problem via novel uncertainty filtering and priority sampling. 
    \item Through comprehensive experiments, we demonstrate that our method consistently outperforms existing approaches across diverse tasks.
\end{itemize}
\section{Related Works}
\subsection{Offline Robust Reinforcement Learning}
The RMDP framework was first introduced by \cite{iyengar2005robust,nilim2005robust} to address the parameter uncertainty problem. The initial works mainly focused on planning problems and have been well-studied \cite{xu2010distributionally,wang2022policy,wang2022convergence}. Recently, robust RL in RMDP has gained much attention, with many works that have studied this problem in online \cite{wang2021online,dong2022online}, and offline \cite{zhou2021finite,shi2022distributionally,ma2022distributionally,blanchet2024double} in both tabular settings \cite{zhou2021finite,yang2022toward,xu2023improved} and large state-action space RMDP setting \cite{ma2022distributionally,panaganti2022robust}. 
We focus on offline robust RL in large state and action spaces. Despite recent successes in such settings \cite{panaganti2022robust}, offline robust RL methods rely on the coverage of the offline dataset. This means their robustness performance, similar to non-robust offline RL algorithms, highly depends on the amount of available offline data \cite{panaganti2022robust}. 
In practice, obtaining datasets with extensive coverage is often infeasible. Therefore, improving the sample efficiency of offline robust RL algorithms is a critical research challenge.

\subsection{Cross-domain Reinforcement Learning}\label{sec:related_works_cross_domain}
Cross-domain RL seeks to improve sample efficiency by leveraging data from additional source environments. 
Domain discrepancies can arise from differences in the observation space \cite{higgins2017darla,bousmalis2018using}, transitional dynamics \cite{eysenbachoff,liudara}.
In this work, we focus on the mismatch in the transition dynamics. Many approaches have been proposed to deal with dynamics mismatches, such as system identification \cite{werbos1989neural,zhu2018fast,chebotar2019closing}, domain randomization \cite{sadeghi2017cad2rl,tobin2017domain,peng2018sim}, or meta-RL \cite{finn2017model,nagabandi2018learning,wu2023zero}. However, these methods often require environment models or careful selection of randomized parameters. Recently, several methods attempted to measure dynamics discrepancy using domain classifiers \cite{eysenbachoff,liudara}, learned dynamics models \cite{liu2024beyond}, or feature representation mismatch \cite{lyucross}. 
Many approaches to utilize source data have been developed, such as reward modification \cite{eysenbachoff,lyucross}, support constraint \cite{liu2024beyond} for purely online \cite{le2024policy,xu2024cross}, purely offline \cite{liu2024beyond,wencontrastive}, or hybrid setting \cite{niu2022trust,niu2023h2o}. 
However, these methods have primarily focused on standard RL setting.
In contrast, we study the \emph{hybrid} cross-domain problem in the \emph{distributionally robust} RL setting, aiming to leverage online source simulators to improve sample efficiency for offline robust RL methods. To this end, we propose a novel approach using uncertainty filtering and priority sampling specifically designed for this hybrid robust setting. 
\subsection{Other robust RL}
Recently, adversarial robust RL \cite{pinto2017robust,rigter2022rambo} and risk-sensitive RL \cite{fei2021exponential,rigter2024one} in online and offline settings also address robustness problems under different frameworks that are independent of RMDP. Additionally, the corruption-robust offline RL problem, where an adversary can modify a fraction of the training dataset, has been studied in \cite{zhang2022corruption,ye2024corruption}. However, their goal is still to find the optimal policy for the nominal model. 
\section{Preliminaries}
We denote an MDP as $\mathcal{M} = (\mathcal{S}, \mathcal{A}, \gamma, r, d_0, P)$, where $\mathcal{S}, \mathcal{A}$ are the state and action spaces. The parameter $\gamma \in (0, 1)$ is the discounted factor, $r: \mathcal{S} \times \mathcal{A} \rightarrow \mathbb{R}$ is the reward function, $d_0$ is the initial state distribution and $P$ is the transition dynamics. We denote a policy $\pi : \mathcal{S} \rightarrow \Delta(\mathcal{A})$ as a map from state space $\mathcal{S}$ to a probability distribution over actions space $\mathcal{A}$. Given a policy $\pi$ and a transition dynamics (model) $P$, we denote discounted state-action occupancy as $d^\pi_P(s, a) = (1 - \gamma)\mathbb{E}_{\pi, P}\left[\sum^\infty_{t=0}\gamma^t\mathds{1}(s_t=s, a_t=a)\right]$. We define value function $V^{\pi, P}$ and state-action value function $Q^{\pi, P}$ for a policy $\pi$ and a transition dynamics $P$ as follows: 
\begin{equation}
    \begin{aligned}
        &V^{\pi, P}(s) = \mathbb{E}_{\pi, P}\big[\sum^\infty_{t=0}\gamma^t r(s_t, a_t)| s_0=s\big], \text{ and } \\
        &Q^{\pi, P}(s, a) = \mathbb{E}_{\pi, P}\big[\sum^\infty_{t=0}\gamma^t r(s_t, a_t)| s_0=s, a_0=a\big].
    \end{aligned}
\end{equation}

We adopt the following notation: $|\mathcal{X}|$ for the cardinality of a set $\mathcal{X}$, $\Delta(\mathcal{X})$ for the set of probability distributions over $\mathcal{X}$, $(x)_+$ for $\max(x, 0)$ where $x \in \mathbb{R}$, and $\mathbb{E}_{P}[f(s')]$ as a short notation for $\mathbb{E}_{s' \sim P(s'|s, a)}[f(s')]$.
\subsection{Distributionally Robust Reinforcement Learning}
We introduce the distributionally robust MDP (RMDP) as $\mathcal{M}_{r} = \{\mathcal{S}, \mathcal{A}, \gamma, r, $ $\mathcal{U}^\sigma_\rho(P^o)\}$. RMDP allows the transition dynamics to be chosen arbitrarily from a predefined uncertainty set $\mathcal{U}^\sigma_{\rho}(P^o)$ centered around a nominal model $P^o$ w.r.t a metric $\rho$. In particular, the uncertainty set is specified as:
 \begin{equation}
 \begin{aligned}
&\mathcal{U}_\rho^\sigma\left(P^o\right):=\otimes \text{ }\mathcal{U}_\rho^\sigma\left(P^o(.|s, a)\right), \\ &\text {with } \mathcal{U}_\rho^\sigma\left(P^o(.|s, a)\right) :=\left\{P(.|s, a) \in \Delta(\mathcal{S}): \rho\left(P(.|s, a), P^o(.|s, a)\right) \leq \sigma\right\},
 \end{aligned}
 \end{equation}
where $\otimes$ denotes the Cartesian product.
In RMDP, we focus on the worst-case performance of a policy $\pi$ over all the transition models in the uncertainty set. Formally, we define the robust value functions for all $s, a \in \mathcal{S}\times \mathcal{A}$ as follows:
\begin{equation}
\begin{aligned}
    V^{\pi, \sigma}(s):=\inf _{P \in \mathcal{U}_\rho^\sigma\left(P^o\right)} V^{\pi, P}(s),   Q^{\pi, \sigma}(s, a):=\inf _{P \in \mathcal{U}_\rho^\sigma\left(P^o\right)} Q^{\pi, P}(s, a).
\end{aligned}
\end{equation}
We also have the following equations held in RMDP:
\begin{equation}
\begin{aligned}
&Q^{\pi, \sigma}(s, a)=r(s, a)+\gamma \inf_{\mathcal{P} \in \mathcal{U}^\sigma_{\rho}\left(P^o\right)} \mathbb{E}_{s'\sim\mathcal{P}} \left[ V^{\pi, \sigma}(s')\right]. \\
\end{aligned}
\end{equation}
In RMDP, there exists at least one deterministic policy that maximizes robust value function \cite{iyengar2005robust}. We denote optimal robust value function, and optimal robust policy, which satisfies the following:
\begin{equation}
\begin{aligned}
    &\forall s \in \mathcal{S}: V^{*, \sigma}(s) := V^{\pi^*, \sigma}(s) = \max_{\pi} V^{\pi, \sigma}, \\
    &\forall s, a \in \mathcal{S} \times \mathcal{A}: Q^{*, \sigma} (s, a) := Q^{\pi^*, \sigma}(s, a) = \max_\pi Q^{\pi, \sigma}(s, a).
\end{aligned}
\end{equation}
Similar to the normal MDP, we have the robust Bellman operator as follows:
\begin{equation}
\begin{aligned}
    &\forall(s, a) \in \mathcal{S} \times \mathcal{A}: \mathcal{T}^\sigma Q(s, a):=r(s, a)+\gamma \inf _{\mathcal{P} \in \mathcal{U}_\rho^\sigma\left(P_{s, a}^0\right)} \mathcal{P} V, \quad V(s):=\max _a Q(s, a).
\end{aligned}\label{eq:rob_bellman_operator}
\end{equation}
It is known that $\mathcal{T}^{\sigma}$ is a contraction mapping w.r.t. the infinity norm, and has a unique fix point solution as $Q^{*, \sigma}$. 
The \emph{fitted} procedure $Q_{k+1} = \mathcal{T}^\sigma Q_k$ can be used to find the fixed point solution $Q^{*, \sigma}$.
\subsection{RMDP with Offline Data}
Offline Robust RL addresses learning robust policies for RMDPs using only an offline nominal dataset $\mathcal{D} = \left\{\left(s_i, a_i, r_i, s'_i\right)\right\}^N_{i=1}$, where $(s_i, a_i) \sim \mu$, $s'_i \sim P^o(.|s_i, a_i)$. The fundamental challenge is that applying the robust Bellman operator in Eq (\ref{eq:rob_bellman_operator}) requires computing expectations over all dynamic models $P \in \mathcal{U}^{\sigma}_{\rho}$,  while only samples from the nominal model $P^o$ are available.

A common approach is to leverage a dual reformulation of the robust Bellman operator, replacing the expectation over all transition dynamics in $\mathcal{U}^{\sigma}_{\rho}(P^o)$ with one over nominal model $P^o$ \cite{panaganti2022robust,shi2022distributionally,ma2022distributionally}. Specifically, \cite{panaganti2022robust} studied uncertainty sets with TV distance and proposed RFQI algorithm. 
To overcome the difficulty of estimating the robust Bellman operator, RFQI proposed the dual reformulation of the second term in the robust Bellman operator.
\begin{restatable}{prop}{dualformtv}\label{prop:dual_form_tv}
    Let $D_{TV}$ be the total variation distance corresponding to the TV uncertainty set $\mathcal{U}^\sigma_{TV}(P^o)$, then
\begin{equation}\label{eq:dual_form_robust_1}
        \begin{aligned}
        \inf_{P\in \mathcal{U}^\sigma_{TV}(P^o)}\mathbb{E}_P[V(s)] = -\inf_{\eta \in [0, \frac{2}{\sigma(1-\gamma)}]}(\mathbb{E}_{P^o}[\left(\eta-V(s)\right)_+] + \sigma((\eta - \inf_{\Tilde{s} \in \mathcal{S}}V(\Tilde{s}))_+ -\eta)
        \end{aligned}
    \end{equation}
\end{restatable}
They made the `fail-state' assumption to overcome the issue of finding $\inf_{s''\in \mathcal{S}}$ $V(s'')$ when $\mathcal{S}$ is large.
\begin{restatable}{assumption}{failstate}\label{assumption:fail_state}
    (Fail-state) The RMDP $\mathcal{M}$ has a `fail-state' $s_f$, such that $\forall a \in \mathcal{A}, \forall P \in \mathcal{U}^\sigma_{TV}(P^o), r(s_f, a) = 0$ and $P(s_f|s_f, a) = 1$.  
\end{restatable}
Under Proposition \ref{prop:dual_form_tv} and Assumption \ref{assumption:fail_state}, RFQI reformulates the robust Bellman operator as follows:
\begin{equation}\label{eq:dual_robust_bellman_operator_simplify}
\begin{aligned}
    &\mathcal{T}^\sigma Q(s, a) := r(s, a) - \gamma \inf _{\eta \in [0, \frac{2}{\sigma(1-\gamma)}]}\big(\mathbb{E}_{s' \sim P^o_{s,a}}[(\eta - V(s'))_+] - \eta(1 - \sigma)\big),
\end{aligned}
\end{equation}
where $\eta \in [0, \frac{2}{\sigma(1-\gamma)}]$ is the dual variable. 
To deal with large state and action space problems, RFQI frames the problem as a function approximation task. Specifically, they learn the dual variable network $g_\theta$ via the loss function $L_{dual} =  \mathbb{E}_{s, a, s' \sim \mathcal{D}}[(g_\theta(s, a) -\max_{a'} Q_\phi(s', a'))_+ - (1-\sigma)g_\theta(s, a)].$
They also define the operator $\mathcal{T}^{\sigma, g}Q(s, a) = r(s, a) - \gamma(\mathbb{E}_{P^o}[(g(s, a) - \max_{a'}Q(s', a'))_{+}]$ $- g(s, a)(1-\sigma))$. Then, the value function $Q_\phi$ is learned with the following objective $L_{RFQI} = \mathbb{E}_{s, a, s' \sim \mathcal{D}}\big[\big(\hat{\mathcal{T}}^{\sigma, g} \hat{Q}_\phi(s, a) - Q_\phi(s, a)\big)^2\big], $
where $\hat{Q}_\phi$ is the value function from the last iteration, and $\hat{\mathcal{T}}^{\sigma, g}$ is the empirical $\mathcal{T}^{\sigma, g}$ that only backs up a single sample.

\section{Hybrid Cross-domain Robust Reinforcement Learning}
\subsection{Problem Setting}
We consider the offline RMDP problem $\mathcal{M}_{r}$ with the uncertainty set around the nominal model $P^o$. 
For clarity, \textbf{we will refer to this nominal model as the target model throughout the paper}.
In our work, we study the RMDPs with total variation (TV) uncertainty set $\mathcal{U}^\sigma_{TV}(P^o)$ and in the large state, action spaces setting. Specifically, we study the setting with limited offline data collected from the target model, i.e. $\mathcal{D} = \left\{\left(s_i, a_i, r_i, s'_i\right)\right\}^N_{i=1}$, where $(s_i, a_i) \sim \mu$, $\mu$ is some data generating distribution, and $s'_i \sim P^o(.|s_i, a_i)$. 
For computational tractability, we adopt the fail-state assumption established in function approximation settings \cite{panaganti2022robust}. We note that the `fail-state' is natural in many real-world systems such as robotics \cite{panaganti2022robust} where the collapse of the robot can be seen as a fail state. 

Along with the offline dataset from the target model, we also have access to an imperfect online simulator which we call a \emph{source environment}. The source environment $\mathcal{M}_{src}$ shares the same state space $\mathcal{S}$, action space $\mathcal{A}$, reward function $r$, discounted factor $\gamma$ and initialized state distribution $d_0$ with target domain, and only differs in its transition model, i.e. $P_{src} \neq P^o$. Our goal is to utilize online simulator $\mathcal{M}_{src}$ and limited offline target dataset $\mathcal{D}$ to learn a policy that is robust under the uncertainty set around target model $P^o$. 
\subsection{Domain Gap for Hybrid Cross-domain Robust RL}\label{sec:domain_gap}
Naively combining source data to train robust target policies can lead to performance degradation due to dynamics mismatch, a challenge also noted in cross-domain RL \cite{niu2022trust,wencontrastive}.
Therefore, caution is necessary when utilizing source data from a different dynamics model. We begin with a theoretical analysis of the performance gap caused by this dynamics mismatch between the two domains, followed by convergence guarantees for value functions in hybrid cross-domain robust RL settings. We provide detailed proof in Appendix 2 due to the space limit.
\begin{restatable}[Performance Bound]{theorem}{performancebound} \label{theorem:performance_bound}
    Let $\mathcal{M}_{src}$ and $\mathcal{M}_{r}$ be the source MDP and the target RMDP with different dynamics $P_{src}$ and $P^o$ respectively. 
    Consider the RMDP with the TV uncertainty set. Denote: $$A = D_{TV}(P^{\pi, \mathcal{U}^\sigma_{TV}(P^o)},  P^{\pi, \mathcal{U}^\sigma_{TV}(\hat{P}^o)}), B = \big|\mathbb{E}_{P_{src}}[V^{\pi, \sigma}_{\hat{P}^o}(s')] - \inf_{P \in \mathcal{U}^{\sigma}(\hat{P}^o)}\mathbb{E}_{P}[V^{\pi, \sigma}_{\hat{P}^o}(s')]\big|$$
    where, given a policy $\pi$, $P^{\pi, \mathcal{U}^\sigma_{TV}(P^o)}$, $P^{\pi, \mathcal{U}^\sigma_{TV}(\hat{P}^o)}$ denote the worst case model w.r.t. the uncertainty set around the target model $P^o$ and the estimated target model $\hat{P^o}$ from offline dataset $\mathcal{D}$, respectively.
    
    The performance difference of any policy $\pi$ on the source domain and the RMDP target can be bounded as follows:
    \begin{equation}\label{eq:transfer_sim_gap}
    \begin{aligned}
    &\mathbb{E}_{s \sim d_0}[V^{\pi, \sigma}(s)] \\&\geq \mathbb{E}_{s \sim d_0}[V^{\pi, src}(s)] -\frac{2\gamma r_{max}}{(1-\gamma)^2}\mathbb{E}_{d^\pi_{P^{\pi, \mathcal{U}^\sigma_{TV}(\hat{P}^o)}}}\big[A\big] - \frac{\gamma}{1 - \gamma}\mathbb{E}_{d^{\pi}_{P_{src}}}\big[B\big].
    \end{aligned}
    \end{equation}
\end{restatable}
The second term $A$ in the Ineq (\ref{eq:transfer_sim_gap}) is caused by the offline dataset and can be reduced by offline Robust RL algorithms via pessimism \cite{panaganti2022robust,shi2022distributionally,blanchet2024double}. 
The third term $B$ represents the gap between the worst case model $P^{\pi, \mathcal{U}^\sigma_{TV}(\hat{P}^o)}$ and the source model $P_{src}$, which can be reduced by our proposed method. 
Specifically, Theorem \ref{theorem:performance_bound} provides the intuition that the robust performance in the target model could be guaranteed if values of robust value function are consistent when evaluating in the source environment and the worst-case model.

Next, we analysis the value function's convergence. Denote the source dataset as $D_{src}$, we consider the following approach using both source and target data:
\begin{equation}\label{eq:Q_update_general}
\begin{aligned}
 Q^{k+1} &\leftarrow \argmin_{Q} \kappa \mathbb{E}_{s, a, s' \sim \mathcal{D}}[(\hat{\mathcal{T}}^{\sigma, g} Q^k - Q)^2]+(1-\kappa) \mathbb{E}_{s, a, s' \sim \mathcal{D}_{src}}[(\hat{\mathcal{T}} Q^k - Q)^2],
\end{aligned}
\end{equation}
where $\kappa \in [0, 1]$ is the combination weight, $k$ denotes training iteration, and $\hat{\mathcal{T}}$ is the empirical Bellman operator. We denote $\mu$ and $\nu$ as the state-action distributions of target and source datasets. We analyze the convergence guarantee of the value function.
To maintain simplicity, we assume the source and the target datasets have the same state-action distribution, i.e. $\mu(s, a) = \nu(s, a), \forall s, a \in \mathcal{S}\times\mathcal{A}$. This assumption can hold easily when the source data $D_{src}$ is generated via a simulator, as it allows flexibility in selecting the transition starting points. We note that the source and target dynamics remain distinct ($P_{src} \neq P^o$). 
Below, we present the convergence guarantee in the following theorem.
\begin{restatable}[Convergence]{theorem}{convergence}
\label{theorem:convergence}
    Let $Q^*$ denote the optimal robust value function for the RMDP of the nominal model $P^o$, and define $Q^0 = 0$. Denote 
    \begin{equation}
    \begin{aligned}
        &\xi = \max_{Q}\max_{s, a \in \mathcal{S}\times\mathcal{A}}\left|\mathcal{T}^{\sigma, g}Q(s, a) - \mathcal{T}^{\sigma}Q(s, a)\right|,\\ 
        &\zeta = \max_{Q}\max_{s, a\in\mathcal{S}\times\mathcal{A}}\left|\mathcal{T}Q(s, a) - \mathcal{T}^{\sigma}Q(s, a)\right|.
    \end{aligned}
    \end{equation}
    Assume $\mu(s, a) = \nu(s, a), \forall s, a \in \mathcal{S}\times\mathcal{A}$, we have the following result holds:
    \begin{equation}
    \label{eq:q_converge_gap}
        \left\| Q^* - Q^{k+1} \right\|_{\infty}\leq \frac{\gamma^{k+1}r_{max}}{1 - \gamma} + \frac{1 - \gamma^{k+1}}{1-\gamma}\big(\kappa\xi + (1-\kappa)\zeta\big).
    \end{equation}
\end{restatable}
Term $\xi$ arises from offline target dataset and can be reduced via offline robust RL algorithms, while $\zeta$ reflects the gap between worst-case target and source models. 
Theorem \ref{theorem:convergence} guarantees that the learned Q function converges near the optimal robust $Q^*$, with a bound determined by the domain gaps and the combination weight. Theorem \ref{theorem:convergence} suggests that target robust performance can be ensured by carefully using the \emph{most} relevant, reliable source data that minimizes domain gaps during training, thus controlling the second term in Ineq (\ref{eq:q_converge_gap}). 

\subsection{Incorporating Source Data in Target Robust Training}
In this section, we present our proposed method, which involves the priority sampling method to select relevant source data for training and uncertainty filtering to keep reliable source samples.

\noindent\textbf{\emph{Source Data Selection with Priority Sampling.}}
Motivated by the performance bound in Theorem \ref{theorem:performance_bound}, we focus on controlling the third term in Ineq (\ref{eq:transfer_sim_gap}). We propose selecting source transitions that induce minimal value discrepancies when incorporating the source environment for training. This requires computing the domain gap between transition pairs starting from the same source state-action pair $(s_{src}, a_{src})$. Specifically, given $(s_{src}, a_{src})$, we aim to estimate the domain gap between next states $s'$, defined as follows:
\begin{equation}
    \begin{aligned}
        \Lambda(s_{src}, a_{src}) = \big|\mathbb{E}_{P_{src}}[V^{\pi, \sigma}_{\hat{P}^o}(s')] - \inf_{P \in \mathcal{U}^{\sigma}(\hat{P}^o)}\mathbb{E}_{P}[V^{\pi, \sigma}_{\hat{P}^o}(s')]\big|.
    \end{aligned}
\end{equation}
Computing $\Lambda(s_{src}, a_{src})$ for a given source state-action pair $(s_{src}, a_{src})$ requires the worst-case model w.r.t. uncertainty set $\mathcal{U}^{\sigma}(\hat{P}^o)$, which is challenging because we only have offline dataset $\mathcal{D}$. However, using Proposition \ref{prop:dual_form_tv} and Assumption \ref{assumption:fail_state},
we can rewrite $\Lambda(s_{src}, a_{src})$ as follows:
\begin{equation} \label{eq:gap_measure}
\begin{aligned} 
    \Lambda(s_{src}, a_{src}) = \big|\mathbb{E}_{P_{src}}[V^{\pi, \sigma}_{\hat{P}^o}(s')] + \inf _{\eta \in [0, \frac{2}{\sigma(1-\gamma)}]}\big(\mathbb{E}_{\hat{P}^o}[(\eta - V^{\pi, \sigma}_{\hat{P}^o}(s'))_+]- \eta(1 - \sigma)\big)\big|.
\end{aligned}
\end{equation}
With Eq (\ref{eq:gap_measure}), given $(s_{src}, a_{src})$ from the source environment, we can approximately compute $\Lambda(s_{src}, a_{src})$ using the offline dataset $\mathcal{D}$.
In practice, we compute $\Lambda(s_{src}, a_{src})$ for each source state-action pair $(s_{src}, a_{src})$ using the robust value function, estimated target model $\hat{P}^o$, and dual variable $\eta$. We learn the estimated dynamics model $\hat{P}^o$ from offline dataset $\mathcal{D}$. For a given $(s_{src}, a_{src})$, we generate the next state $s'_{tar}$ using the learned target dynamics model. The learned dual network $g_\theta$ gives approximated values for dual variable $\eta$. We then introduce a gap-measurement function approximating $\Lambda(s_{src}, a_{src})$ by estimating the value function for next state $s'$ as $V^{\pi, \sigma}_{\hat{P}^o}(s') := Q_{\phi}(s', a')|_{a' \sim \pi(.|s')}$:
\begin{equation}\label{eq:score_function}
     \begin{aligned}
        \hat{\Lambda}(s_{src}, a_{src}) &= \big|Q_\phi(s'_{src}, a') +   (g_\theta(s_{src}, a_{src})-Q_\phi(s'_{tar}, a'))_+ \\ &- g_\theta(s_{src}, a_{src})(1-\sigma)\big|.
     \end{aligned}
 \end{equation}
As our objective is to select the source samples with small gaps to tighten the bound in Ineq (\ref{eq:transfer_sim_gap}), we introduce the following \emph{priority score function} $\psi(s, a) = 1/(1 + \hat{\Lambda}(s, a)).$
This priority score guides our sampling process during training. The sampling probability for each source transition is defined as:
\begin{equation}\label{eq:probability_prb}
    p^i(s, a, s', r) = \psi^i(s, a)/\sum_k \psi^k(s, a),
\end{equation}
where $\psi^i(s, a)$ is the priority score of the source transition.

\noindent\textbf{\emph{Uncertainty Filtering.}}\label{sec:algorithm_unc_filter}
To address the uncertainty of the offline dataset, we employ a quantifier to compute uncertainty values for each source sample. Source samples with high uncertainty can lead to unreliable estimation scores, potentially hindering the training process if included. Conversely, reliable source state-action pairs with low uncertainty can serve as valuable augmented data for training the dual network $g_\theta$.
Therefore, we propose removing source samples with high uncertainty values. Inspired by prior works \cite{kidambi2020morel,wu2024ocean}, we train an ensemble of $N$ dynamics model $\{\hat{P}^o_i(s'|s, a) = \mathcal{N}\left(\mu_{\varphi}(s, a), \Sigma_{\varphi}(s, a)\right)\}^N_{i=1}$. Each model in the ensemble is trained using offline target dataset $\mathcal{D}_{tar}$ via the maximum log-likelihood (MLE) as follows: $\mathcal{L}_{\varphi}=\mathbb{E}_{\left(s, a, r, s'\right) \sim \mathcal{D}}[\log \hat{P}^o(s' \mid s, a)].$

Then, we use the max pairwise difference as our uncertainty quantifier, i.e. $u(s, a) = \max_{i, j}\|\mu^i_{\varphi}(s, a) - \mu^j_{\varphi}(s, a)\|^2$, where $\|.\|_2$ is the L2-norm and $\mu^i_{\varphi}(s, a)$, $\mu^j_{\varphi}(s, a), i, j \in \{1, \dots, N\}$ are the mean vectors of the Gaussian distributions in the ensemble dynamics model. For the uncertainty threshold, instead of naively setting a constant threshold, we measure the uncertainty on all samples in the offline dataset $\mathcal{D}$. Then, we take the maximum uncertainty in the dataset and set the uncertainty threshold as follows: $\epsilon_u = \frac{1}{\alpha}\max_{s, a \in \mathcal{D}}u(s, a),$
where $\alpha \in \mathbb{R}_+$ is a hyperparameter to control the threshold value.  
Then, for any source transition $(s, a, s', r)_{src}$, we add them to the source replay buffer if its uncertainty value is less than the uncertainty threshold $\epsilon_u$. Otherwise, we remove them.


During training, we sample a batch of source data from the source replay buffer with probabilities defined by Eq. (\ref{eq:probability_prb}). Based on Theorems \ref{theorem:performance_bound} and \ref{theorem:convergence}, we prioritize the most relevant source samples while carefully controlling combination weight $\kappa$. We recompute scores for these samples, select the top-k highest-scoring samples to update the value function, and adjust priorities accordingly. The robust value function is trained using both source and target data as follows:
\begin{equation}\label{eq:hydro_Q_update}
\begin{aligned}
 Q_\phi \leftarrow \argmin_{Q_\phi} \mathbb{E}_{s, a, s' \sim \mathcal{D}}[(\hat{\mathcal{T}}^{\sigma, g} \hat{Q}_\phi - Q_\phi)^2]+ \mathbb{E}_{s, a, s' \sim \mathcal{D}_{src}}[\omega(s, a, s^{\prime})(\hat{\mathcal{T}} \hat{Q}_\phi - Q_\phi)^2]
\end{aligned}
\end{equation}
, where source data sampled via priority sampling after the uncertainty filter, $w(s, a, s') = \mathds{1}(\psi(s, a) > \psi_{k\%})$ , and $\mathcal{T}$ and $\mathcal{T}^\sigma$ is the normal and robust Bellman operator respectively.
We use the offline target data $(s_{tar}, a_{tar}, s'_{tar})$ along with the augmented sample $(s_{src}, a_{src}, s'_{tar})$, where ${s'_{tar} \sim \hat{P}^o(s'|s_{src}, a_{src})}$, for training dual network $g_\theta$. Specifically, we update the dual network $g_\theta$ as follows:
\begin{equation}\label{eq:hydro_g_update}
\begin{aligned}
    \theta &\leftarrow \argmin_\theta \mathbb{E}_{s, a, s' \sim \mathcal{D}}\left[\left(g_\theta(s, a) - V(s')\right)_+ - (1-\sigma)g_\theta(s, a)\right] \\&+ \mathbb{E}_{s, a \sim \mathcal{D}_{src}, s' \sim \hat{P}^0}\left[\left(g_\theta(s, a) - V(s')\right)_+ - (1-\sigma)g_\theta(s, a)\right].
\end{aligned}
\end{equation}
\begin{algorithm}[t]
\caption{HYbrid cross-Domain RObust RL - HYDRO}
\label{alg:robust}
\begin{algorithmic}[1]\label{alg:proposed}
\STATE\textbf{Input:} Source $\mathcal{M}_{src}$, offline target dataset $\mathcal{D}$, the source replay buffer $\mathcal{D}_{src} = \emptyset$, robust value function $Q_\phi$ and dual variable functions $g_\theta$.
\STATE Train $\{\hat{P}^o_i(s'|s, a) = \mathcal{N}\left(\mu_{\varphi}(s, a), \Sigma_{\varphi}(s, a)\right)\}^N_{i=1}\}$ via MLE on $\mathcal{D}$.
\FOR {$t = 1, \dots, \text{num iterations}$} 
    \FOR{$i=1, \dots, h$} 
        \STATE Rollout with $\mathcal{M}_{src}$, compute $u_i(s_i, a_i) = \max_{j, k}\|\mu^j_{\varphi}(s_i, a_i) - \mu^k_{\varphi}(s_i, a_i)\|^2$.
        \IF{$u_i \leq \epsilon_u$} 
            \STATE $\mathcal{D}_{src} \leftarrow \mathcal{D}_{src} \bigcup (s_i, a_i, r_i, s'_i)$. 
        \ENDIF
    \ENDFOR
    \STATE Sample $\{(s, a, r, s')^i_{src}\}^{N}_{i=1}$ with probability $p^i(s, a, s')$ via Eq (\ref{eq:probability_prb}) from $\mathcal{D}_{src}$. 
    \STATE Sample $\{(s, a, r, s')^i_{tar}\}^{N}_{i=1}$ uniformly from $\mathcal{D}$.
    \STATE Update transition priority in $\{(s, a, r, s')^i_{src}\}^{N}_{i=1}$.
    \STATE Update $g_\theta$ via Eq (\ref{eq:hydro_g_update}) using source, target data.
    \STATE Update $Q_\phi$ via Eq (\ref{eq:hydro_Q_update}) using $\{(s, a, r, s')^i_{src}\}^{N}_{i=1}$ and $\{(s, a, r, s')^i_{tar}\}^{N}_{i=1}$.
\ENDFOR
\STATE\textbf{return } $Q_\phi$.
\end{algorithmic}
\end{algorithm}
\noindent\textbf{\emph{Algorithm.}}
We summarize the above steps as our proposed method HYDRO in Algorithm \ref{alg:robust}.
\section{Experiments}
In this section, we present the empirical evaluation to answer the following questions: \textbf{1)} Can HYDRO enhance data efficiency and improve robustness performance in scarce data settings? \textbf{2)}  Why is using HYDRO more advantageous than just naively merging the source data? \textbf{3)} How do different components of HYDRO contribute to its performance?

\noindent\textbf{\emph{Environments.}}
We conduct our experiments on three MuJoCo environments (HalfCheetah-v3, Walker2d-v3, Hopper-v3), utilizing the Medium datasets from D4RL \cite{fu2020d4rl} as our offline datasets \cite{panaganti2022robust}. To create the scarce data settings, we only use 10\% of these datasets for training, i.e. 100K target transitions from D4RL. The source environments are created by modifying the morphology of the agents in the Mujoco XML file. We consider two types of modifications: single-comp, modifying a single agent component, and multi-comp, altering multiple components. For robustness evaluation, we perturb each Mujoco environment by altering its physical parameters. 

\noindent\textbf{\emph{Baselines}.}
We compare HYDRO against the following baselines: \emph{RQFI} \cite{panaganti2022robust}, the current state-of-the-art in offline robust RL; \emph{H2O} \cite{niu2022trust}, an only recent state-of-the-art method with available public code for non-robust hybrid cross-domain transfer that uses importance sampling to correct the dynamics shift between source and target environments; \emph{PQL} \cite{liu2020provably}, a non-robust offline RL algorithm and a practical state-of-the-art variant of FQI with neural architecture. Finally, we train RFQI agent with a full offline target dataset, which we refer to as \emph{Oracle}.
Due to limited space, we defer more details about the environment and baseline settings to Appendix 4 and provide more experiment results in Appendix 5.
\subsection{Robustness Performance Evaluation}\label{sec:robust_performance_comparison}
\begin{figure*}[t]
    \centering
    \includegraphics[width=0.85\linewidth]{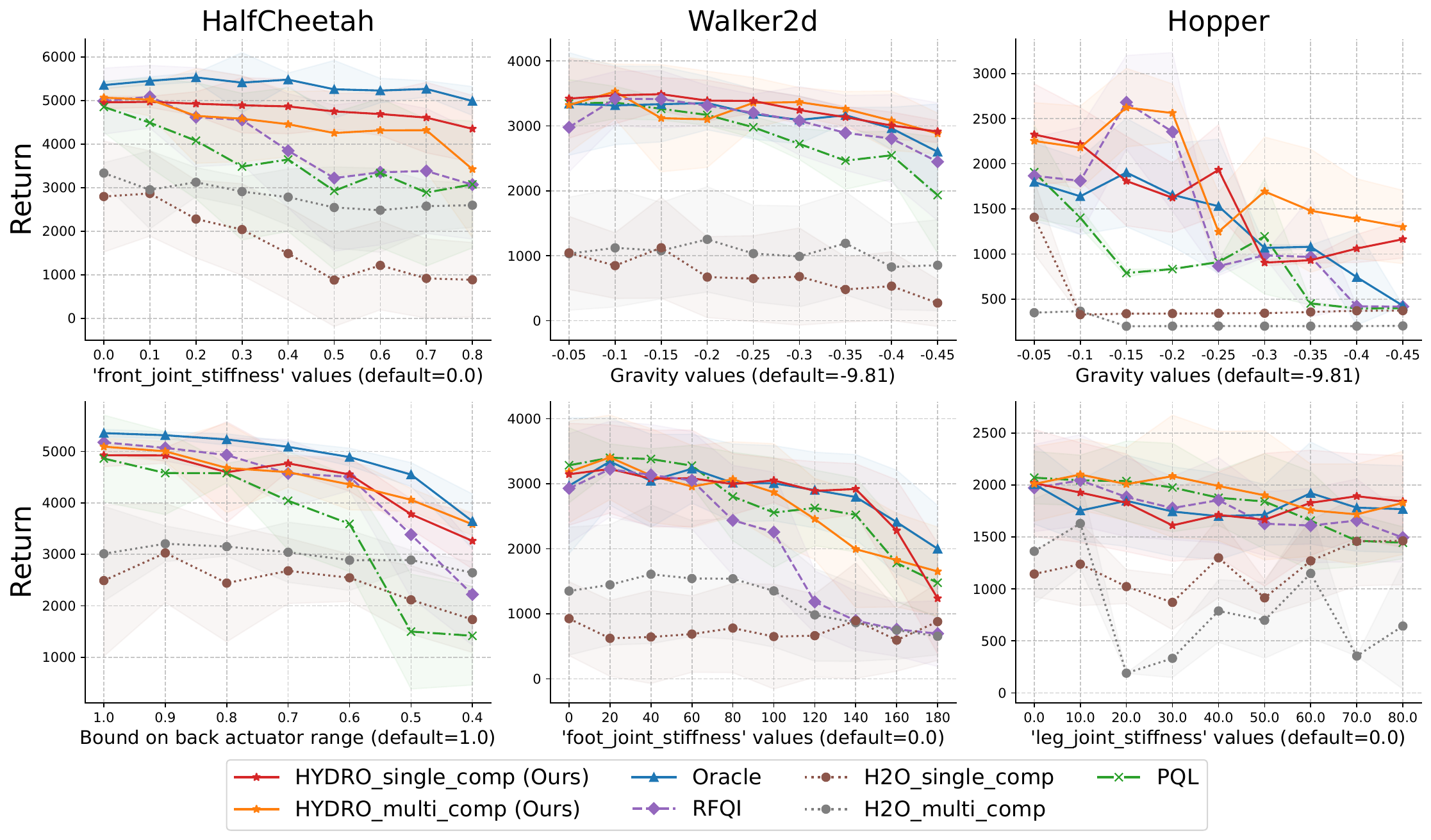}
    \caption{Cumulative rewards of different methods in three Mujoco benchmarks under perturbation. The lines are the average returns over 30 different seeded runs, and the shaded areas represent standard deviation.}
    \label{fig:main_exps}
\end{figure*}
\begin{table*}[t]
\centering
\caption{Average returns over different environment parameter perturbations for Mujoco tasks over 30 different seeded runs. We \textbf{bold} the best results (except Oracle). ``-m": multi comp, ``-s" single comp, ``B-" Back. }
\label{tab:main}
\begin{tabular}{lcccccc}
\toprule
\multirow{2}{*}{}        & \multicolumn{2}{c}{Halfcheetah}                             & \multicolumn{2}{c}{Walker2d}                     & \multicolumn{2}{c}{Hopper}                               \\  
                         &\makecell{Front joint \\ stiffness}   & \makecell{B-actuator \\ ctrlrange}  & Gravity        & \makecell{Foot joint \\ stiffness} & Gravity                  & \makecell{Leg joint \\ stiffness} \\ \cmidrule(){1-7}
Oracle                   & 5332±287         & 4866±184           &  3150±481   & 2871±684        & 1316±373                   & 1804±398      \\
\cmidrule(){1-7}
RFQI                     & 4016±1219                  & 4264±557                   & 3062±537          & 2056±672                &  1374±358            & 1769±361                \\
PQL                      & 3644±1437                  & 3508±828                   & 2865±340          & 2710±506                 & 920±151                    &  1823±323         \\
H2O-m           & 2814±606                  & 2974±350                    & 1043±778          & 1208±770                 & 235±3                     & 794±334                 \\
H2O-s          & 1710±1013                  & 2431±854                    & 701±585           & 733±655                  & 466±47                    & 1187±316                \\
\cmidrule(){1-7}
HYDRO-m &  4456±872            & \textbf{ 4480±350}              & 3225±487    &  2652±703           & \textbf{1858±468} & \textbf{1933±468}       \\
HYDRO-s & \textbf{ 4781±476}            & 4399±382              & \textbf{3273±284} &  \textbf{2790±623}           &  1551±309             & 1814±469          \\
\bottomrule 
\end{tabular}
\end{table*}
In this section, we answer the first question, showing that our method, HYDRO, can enhance data efficiency and improve robust performance in scarce data settings.
Figure \ref{fig:main_exps} presents the performance of our method and the baselines across three Mujoco environments under model parameter perturbations. Notably, the robust performance of RFQI degrades substantially with reduced training data. In the scarce data setting (10\% target data), RFQI's performance drops notably with increasing perturbations, resembling the non-robust offline method PQL. H2O performs poorly across all settings, as its performance is heavily reliant on the amount of training target data and struggles in scarce data scenarios \cite{niu2022trust}. We believe the lack of target samples causes the inferior performance of H2O, which also was observed in \cite{daoudi2024conservative}. 
In contrast, HYDRO consistently demonstrates robust performance, surpassing RFQI and non-robust methods across all tasks. 
While baseline methods exhibit substantial performance drops with increasing environmental changes, HYDRO maintains robustness.
Table \ref{tab:main} presents the average returns of all methods under various environment parameter perturbations. As the table illustrates, our method improves upon RFQI across all tasks, with the most significant improvement reaching approximately 36\%. 
Statistical testing (please see Appendix 5.1) confirms HYDRO significantly outperforms all baselines.
Importantly, compared to Oracle, HYDRO exhibits the smallest degradation in robust performance across all tasks.
\subsection{Ablation study.}\label{sec:exp_ablation}
\begin{figure*}[t]
        \begin{subfigure}[b]{0.32\textwidth}
                \includegraphics[width=1\textwidth]{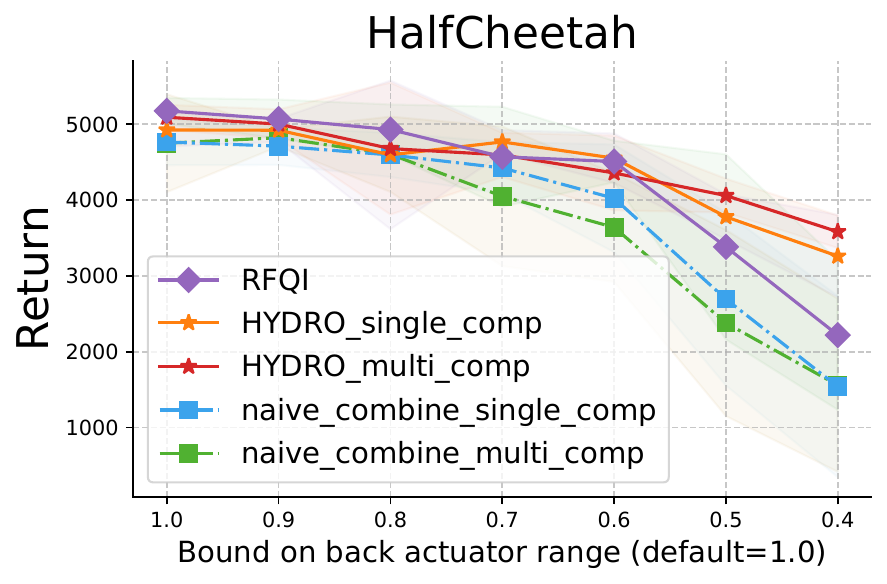}
                \caption{Naively combining data.}
                \label{fig:naive_combining}
        \end{subfigure}%
        \begin{subfigure}[b]{0.32\textwidth}
                \includegraphics[width=1\textwidth]{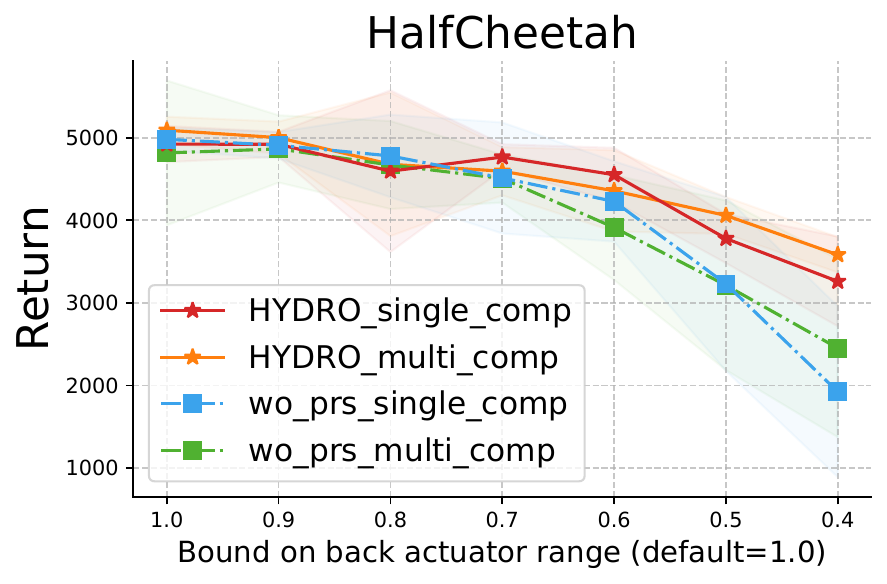}
                \caption{Priority sampling.}
                \label{fig:ablation_prb}
        \end{subfigure}%
        \begin{subfigure}[b]{0.32\textwidth}
                \includegraphics[width=1\textwidth]{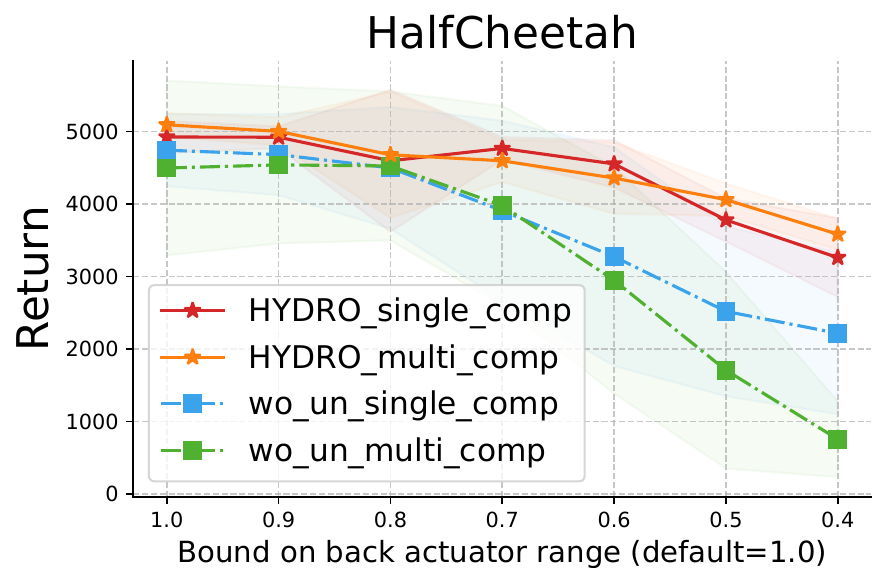}
                \caption{Uncertainty filter.}
                \label{fig:ablation_unc}
        \end{subfigure}%
        \caption{(a) Robust performance comparison between HYDRO, RFQI, and its variations using naive combination of source and target data. (b-c) Robust performance comparison between HYDRO and its variants without priority sampling and uncertainty filter.}
\end{figure*}

\noindent\textbf{\emph{Naively combining source data.}}\label{sec:ablation_naive}
To address the second question, we compare the performance of RFQI trained on target data only versus RFQI trained on combined target and source data (cross-domain data), as well as our proposed method. For the cross-domain data experiment, we simply merge target and source data without any further processing and use this combined dataset to train RFQI.
Figure \ref{fig:naive_combining} demonstrates that simply incorporating additional source data does not enhance robustness and can also lead to poor performance compared to using only the limited target data (100K). We argue that the primary reason for this is the dynamics mismatch between the source and worst-case model, which hinders the simple merging cross-domain data approach. On the other hand, our method handles this mismatch by selecting the \emph{most} reliable source data with a small gap to tighten Ineq (\ref{eq:transfer_sim_gap}) and control the learn robust Q function as motivated by Theorem \ref{theorem:convergence}. The results highlight the effectiveness of our approach compared to both RFQI and the naive combination strategy.

To address the third question, we perform a comprehensive ablation study on HYDRO, analyzing the contribution of each component to its performance.

\begin{figure}[t]
    \centering
    \includegraphics[width=0.4\linewidth]{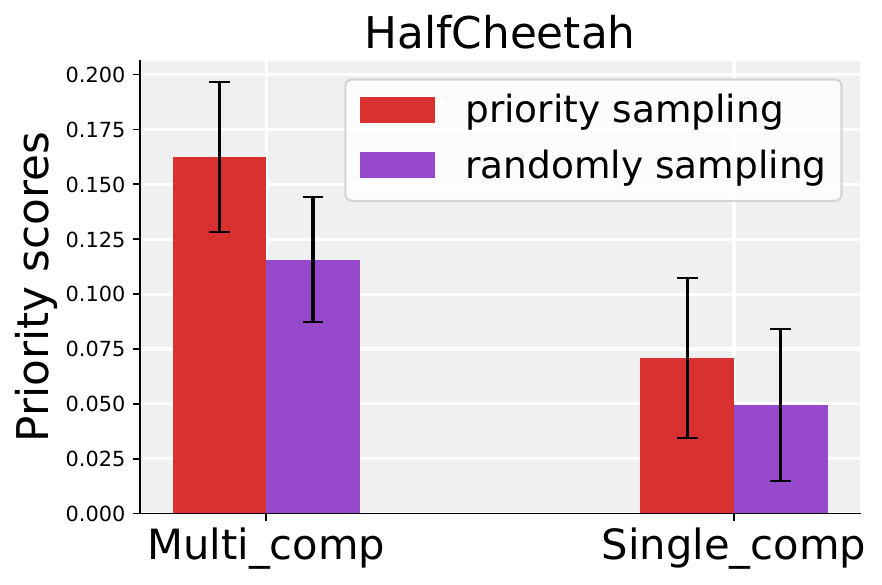}
    \caption{Average priority scores of random and priority sampling.}
    \label{fig:priority_score}
\end{figure}

\noindent\textbf{\emph{Priority Sampling.}}\label{sec:ablation_priority} 
To evaluate priority sampling's impact, we compare against a variant without this component. 
Figure \ref{fig:ablation_prb} shows a significant decrease in robust performance when priority sampling is excluded.
The enhanced performance stems from the increased utilization of source samples with greater proximity to the worst-case target model. Figure \ref{fig:priority_score} confirms this hypothesis, demonstrating that priority sampling significantly increases the mean priority score of selected samples compared to random sampling. 
These results confirm that priority sampling plays a crucial role in enhancing the robustness of our method.

\noindent\textbf{\emph{Uncertainty Filter.}}\label{sec:ablation_unc} To assess the impact of the uncertainty filter, we compare our method to a variant that excludes this component. Figure \ref{fig:ablation_unc} reveals a significant decrease in robustness when the uncertainty filter is omitted, highlighting its crucial role in our approach. We hypothesize that the significant performance gains observed when incorporating the uncertainty filter arise from the strategic exclusion of source samples with high uncertainty levels, allowing the model to focus on more informative data points, leading to improved performance.

\subsection{More Experiment Results}
We conduct additional experiments to explore the potential of HYDRO. Please see Appendix 5 for more results.

\noindent\textbf{\emph{HYDRO with different domain-gaps measurement.}}
To emphasize the distinction between HYDRO and \emph{standard} cross-domain RL methods, we evaluate HYDRO's performance using an alternative domain-gaps measurement. We replace our measurement, which quantifies the discrepancy between worst-case target and source models, with standard cross-domain RL measurements quantifying the discrepancy between nominal target and source models using domain classifiers from DARC \cite{eysenbachoff}. 
Results in Appendix 5.4 illustrate this substitution leads to significant performance degradation. These results underscore the inadequacy of standard domain-gap measurement in \emph{robust} RL settings and highlight the critical role and effectiveness of HYDRO's tailored measurement approach.

\noindent\textbf{\emph{How HYDRO performs under harder limited target data settings?}}\label{sec:discussion_harder}
To further understand H2O's performance, we analyze HYDRO's behavior under increasingly challenging, data-limited target settings. 
The results in Appendix 5.5 show RFQI's robust performance decreases substantially as target data decreases, while HYDRO maintains consistently strong performance with only minimal degradation. These results demonstrate HYDRO's effectiveness in overcoming data scarcity challenges.


\section{Conclusion}\label{sec:conclusion}
In this paper, we have addressed the problem of Hybrid cross-domain robust reinforcement learning, which is widely encountered in many real-world problems. To the best of our knowledge, this is the first work to tackle the hybrid setting of online source and offline target under Robust MDPs. We introduce HYDRO, a novel method that effectively leverages source domain data by selecting relevant and reliable data points with respect to the worst-case model in an uncertainty set, utilizing priority sampling and an uncertainty filter. We have demonstrated the superior performance of our method through extensive experiments. The limitation of our approach is its dependence on an estimated target transition model. Although we have demonstrated our method's effectiveness empirically, a theoretical analysis of how the estimated target model impacts the performance of the learned policy can be a promising direction for future research.

\newpage
\section*{APPENDIX}
\setcounter{section}{0}  
\renewcommand{\thesection}{\arabic{section}}  
\section{Extended Related Works}
\subsection{Cross-domain Reinforcement Learning}\label{sec:related_works_cross_domain}
Cross-domain RL seeks to improve sample efficiency by leveraging data from additional source environments. 
Domain discrepancies can stem from differences in the observation space \cite{higgins2017darla,bousmalis2018using}, transitional dynamics \cite{eysenbachoff,liudara}, agent embodiment \cite{liu2022revolver}, etc.
In this work, we specifically focus on the mismatch in the transition dynamics. Many approaches have been proposed to deal with dynamics mismatches, such as system identification \cite{werbos1989neural,zhu2018fast,chebotar2019closing}, which aligns the source dynamics with the observed target data, domain randomization \cite{sadeghi2017cad2rl,tobin2017domain,peng2018sim}, which trains the policies over a collection of randomized simulated source domains, or meta-RL \cite{finn2017model,nagabandi2018learning,wu2023zero}. However, these methods often require environment models or careful selection of randomized parameters. Recently, several methods attempted to measure dynamics discrepancy using domain classifiers \cite{eysenbachoff,liudara}, learned dynamics models \cite{liu2024beyond}, mutual information \cite{wencontrastive}, or feature representation mismatch \cite{lyucross}.  
Many approaches to utilize source data have been developed, such as reward modification \cite{eysenbachoff,lyucross}, support constraint \cite{liu2024beyond} for purely online \cite{le2024policy,xu2024cross}, purely offline \cite{liu2024beyond,wencontrastive}, or hybrid setting \cite{niu2022trust,niu2023h2o}. 
However, these methods have primarily focused on standard RL settings, which aim to improve the policy performance in the target domain.
In contrast, we study the \emph{hybrid} cross-domain problem in the \emph{distributionally robust} RL setting, focusing on improving the \emph{target robust performance} w.r.t. the uncertainty set around the target model. Our work aims to leverage online source simulators to improve sample efficiency for offline robust RL methods. To this end, we propose a novel approach using uncertainty filtering and priority sampling specifically designed for this hybrid robust setting. 
\section{Missing Theoretical Proofs}\label{sec:proofs}
In this section, we provide formal proofs omitted from the main paper. 
\subsection{Dual Formulation for Robust Bellman Operator}
Let denote the $f$-divergence between two distributions $P$ and $P^o$ as follows:
\begin{equation}
    D_f(P||P^o) = \int f(\frac{\mathrm{d}P}{\mathrm{d}P^o}) \mathrm{d}P^o,
\end{equation}
where $f$ is a convex function. We have different divergences for different forms of the function $f$, such as $f(t) = |t-1|/2$ gives Total variation (TV), or $f(t)=t\log t$ gives the KL divergence, etc.  
We have the following result from distributional robust optimization \cite{shapiro2017distributionally,duchi2021learning}.
\
\begin{lemma}\label{lemma:dual_lemma}
Let $D_f$ be the f-divergence, then we have
\begin{equation}
\sup _{D_f\left(P \| P^o\right) \leq \sigma} \mathbb{E}_P[V(s')]=\inf _{\lambda>0, \eta \in \mathbb{R}} \mathbb{E}_{P^o}\left[\lambda f^*\left(\frac{V(s')-\eta}{\lambda}\right)\right]+\lambda \sigma+\eta,
\end{equation}
where, $P^o$ is a distribution on the space $\mathcal{X}$, $l:\mathcal{X} \rightarrow \mathbb{R}$ is a function mapping from $\mathcal{X}$ to $\mathbb{R}$, and $f^*(s)=\sup_{t\geq 0}\{st-f(t)\}$ is the Fenchel conjugate.
\end{lemma}

\emph{Proof.} We refer to \cite{shapiro2017distributionally} (Section 3.2) for the details of the proof.

When we choose the function $f(t) = |t-1|/2$, which corresponds to the TV metric, we have the following result.
\begin{restatable}{prop}{dualformtv}\label{prop:dual_form_tv}
    Let $D_{TV}$ be the total variation distance corresponding to the TV uncertainty set $\mathcal{U}^\sigma_{TV}(P^o)$, then
\begin{equation}\label{eq:dual_form_robust_1}
        \begin{aligned}
        \inf_{P\in \mathcal{U}^\sigma_{TV}(P^o)}\mathbb{E}_P[V(s)] = -\inf_{\eta \in [0, \frac{2}{\sigma(1-\gamma)}]}(\mathbb{E}_{P^o}[\left(\eta-V(s)\right)_+] + \sigma((\eta - \inf_{\Tilde{s} \in \mathcal{S}}V(\Tilde{s}))_+ -\eta)
        \end{aligned}
    \end{equation}
\end{restatable}

\emph{Proof.} This is Lemma 5 in \cite{panaganti2022robust}. Here, we provide a brief proof. For the detailed version, please check \cite{panaganti2022robust}.

First, we compute the Fenchel conjugate of $f(t) = |t-1|/2$, as follows:
\begin{equation}
\begin{aligned}
    f^*(s) &=\sup _{t \geq 0}\left\{s t-\frac{1}{2}|t-1|\right\}
    \\
    &=\max \left\{\sup _{t \in[0,1]}\left\{\left(s+\frac{1}{2}\right) t-\frac{1}{2}\right\}, \sup _{t>1}\left\{\left(s-\frac{1}{2}\right) t+\frac{1}{2}\right\}\right\}
    \\
    &=\begin{cases}-\frac{1}{2} & s \leq-\frac{1}{2} \\ s & s \in\left[-\frac{1}{2}, \frac{1}{2}\right]. \\ +\infty & s>\frac{1}{2}\end{cases}
\end{aligned}
\end{equation}
Then, from Lemma \ref{lemma:dual_lemma}, we have
\begin{equation}
\begin{aligned}
&\sup _{D_f\left(P \| P^o\right) \leq \sigma} \mathbb{E}_P[V(s')] \\ & =\inf _{\lambda>0, \eta \in \mathbb{R}} \mathbb{E}_{P^o}\left[\lambda f^*\left(\frac{V(s')-\eta}{\lambda}\right)\right]+\lambda \sigma+\eta \\
& =\inf _{\lambda, \eta: \lambda>0, \eta \in \mathbb{R}, \frac{\sup _{s' \in \mathcal{S}} V(s')-\eta}{\lambda} \leq \frac{1}{2}} \mathbb{E}_{P^o}\left[\lambda \max \left\{\frac{V(s')-\eta}{\lambda},-\frac{1}{2}\right\}\right]+\lambda \sigma+\eta
\\
&= \inf _{\lambda, \eta: \lambda>0, \eta \in \mathbb{R}, \frac{\sup _{s' \in \mathcal{S}} V(s')-\eta}{\lambda} \leq \frac{1}{2}} \mathbb{E}_{P o}\left[(V(s')-\eta+\lambda / 2)_{+}\right]-\lambda / 2+\lambda \sigma+\eta
\\
&= \inf _{\lambda, \eta^{\prime}: \lambda>0, \eta^{\prime} \in \mathbb{R}, \frac{\sup _{s' \in \mathcal{S}} V(s')-\eta^{\prime}}{\lambda} \leq 1} \mathbb{E}_{P o}\left[\left(V(s')-\eta^{\prime}\right)_{+}\right]+\lambda \sigma+\eta^{\prime}.
\end{aligned}
\end{equation}
The second equality comes from $f^*(\frac{V(s') - \eta}{\lambda}) = +\infty$ whenever $\frac{V(s') - \eta}{\lambda} > 1/2$, which can be ignored as we are minimizing over $\eta, \lambda$. The third equality follows that $\max(x, y) = (x-y)_+ +y$ for any $x, y \in \mathbb{R}$. The last equality follows by making $\eta' = \eta - \lambda/2$. We take the optimal value of $\lambda = (\sup_{s''\in\mathcal{S}}(V(s') - \eta')_+$, we have the following:
\begin{equation}
\begin{aligned}
    \sup _{D_f\left(P \| P^o\right) \leq \sigma} \mathbb{E}_P[V(s')]=\inf _{\eta \in \mathbb{R}} \left(\mathbb{E}_{P^{\circ}}\left[(V(s')-\eta)_{+}\right]+(\sup _{s'' \in \mathcal{S}} V(s'')-\eta)_{+} \sigma+\eta \right) .
\end{aligned}
\end{equation}
Finally, we have:
\begin{equation}
\begin{aligned}
    &\inf _{D_f\left(P \| P^o\right) \leq \sigma} \mathbb{E}_P[V(s')] \\& =-\sup _{D_f\left(P \| P^o\right) \leq \sigma} \mathbb{E}_P[-V(s')] \\
    & =-\inf _{\eta \in \mathbb{R}} \left(\mathbb{E}_{P^o}\left[(-V(s')-\eta)_{+}\right]+(\sup _{s'' \in \mathcal{S}}-V(s'')-\eta)_{+} \sigma+\eta \right) \\
    & =-\inf _{\eta^{\prime} \in \mathbb{R}} \left(\mathbb{E}_{P^o}\left[\left(\eta^{\prime}-V(s')\right)_{+}\right]+(\eta^{\prime}-\inf _{s'' \in \mathcal{S}} V(s''))_{+} \sigma-\eta^{\prime}\right),
\end{aligned}
\end{equation}
which completes the proof.
 
\subsection{Domain Gap for Hybrid Cross-domain Robust RL}\label{sec:appendix_domain_gaps}

We provide a theoretical guarantee for using additional source data to improve the performance of the RMDP in the target model under dynamics mismatch. We have the following performance bound for any policy in hybrid cross-domain robust RL:
\begin{restatable}[Performance Bound]{theorem}{performancebound} \label{theorem:performance_bound}
    Let $\mathcal{M}_{src}$ and $\mathcal{M}_{r}$ be the source MDP and the target RMDP with different dynamics $P_{src}$ and $P^o$ respectively. 
    Consider the RMDP with the TV uncertainty set. The performance difference of any policy $\pi$ on the source domain and the RMDP target can be bounded as follows:
    \begin{equation}\label{eq:transfer_sim_gap}
    \begin{aligned}
    &\mathbb{E}_{s \sim d_0}[V^{\pi, \sigma}(s)] \\&\geq \mathbb{E}_{s \sim d_0}[V^{\pi, src}(s)] -\frac{2\gamma r_{max}}{(1-\gamma)^2}\mathbb{E}_{d^\pi_{P^{\pi, \mathcal{U}^\sigma_{TV}(\hat{P}^o)}}}\big[D_{TV}\big(P^{\pi, \mathcal{U}^\sigma_{TV}(P^o)},  P^{\pi, \mathcal{U}^\sigma_{TV}(\hat{P}^o)}\big)\big] \\&- \frac{\gamma}{1 - \gamma}\mathbb{E}_{d^{\pi}_{P_{src}}}\big[\big|\mathbb{E}_{P_{src}}[V^{\pi, \sigma}_{\hat{P}^o}(s')] - \inf_{P \in \mathcal{U}^{\sigma}(\hat{P}^o)}\mathbb{E}_{P}[V^{\pi, \sigma}_{\hat{P}^o}(s')]\big|\big],
    \end{aligned}
    \end{equation}
    where, given a policy $\pi$, $P^{\pi, \mathcal{U}^\sigma_{TV}(P^o)}$ denotes the worst case model w.r.t. the uncertainty set around the target model $P^o$, $P^{\pi, \mathcal{U}^\sigma_{TV}(\hat{P}^o)}, V^{\pi, \sigma}_{\hat{P}^o}(s')$ denotes worst case model and robust value function w.r.t. the uncertainty set around estimated target model $\hat{P}^o$ from offline dataset $\mathcal{D}$.
\end{restatable}
\emph{Proof}: Given a police $\pi$, denote the worst-case model w.r.t. the uncertainty set around the target model $P^o$ is $P^{\pi, \mathcal{U}^\sigma_{TV}(P^o)}$, and the worst-case model w.r.t. the uncertainty set around the estimated target model $\hat{P}^o$ is $P^{\pi, \mathcal{U}^\sigma_{TV}(\hat{P}^o)}$.
For the performance bound for any policy $\pi$, we first convert it to the following form:
\begin{equation}
\begin{aligned}
    &\mathbb{E}_{s \sim d_0}[V^{\pi, \sigma}(s)] - \mathbb{E}_{s \sim d_0}[V^{\pi, src}(s)] \\&= \underbrace{\mathbb{E}_{s \sim d_0}[V^{\pi, \sigma}(s)] - \mathbb{E}_{s \sim d_0}[V^{\pi, \sigma}_{\hat{P}^o}(s)]}_{(a)} + \underbrace{\mathbb{E}_{s \sim d_0}[V^{\pi, \sigma}_{\hat{P}^o}(s)] - \mathbb{E}_{s \sim d_0}[V^{\pi, src}(s)]}_{(b)}.
\end{aligned}
\end{equation}
For term $(a)$ in the RHS, based on the telescoping lemma \cite{luoalgorithmic}, we have:
\begin{equation}
\begin{aligned}
    &\mathbb{E}_{s \sim d_0}[V^{\pi, \sigma}(s)] - \mathbb{E}_{s \sim d_0}[V^{\pi, \sigma}_{\hat{P}^o}(s)] \\&= \frac{\gamma}{1 - \gamma}\mathbb{E}_{d^\pi_{P^{\pi, \mathcal{U}^\sigma_{TV}(\hat{P}^o)}}}\left[\mathbb{E}_{P^{\pi, \mathcal{U}^\sigma_{TV}(P^o)}}[V^{\pi, \sigma}(s')] - \mathbb{E}_{P^{\pi, \mathcal{U}^\sigma_{TV}(\hat{P}^o)}}[V^{\pi, \sigma}(s')] \right] \\
    &= \frac{\gamma}{1 - \gamma}\mathbb{E}_{d^\pi_{P^{\pi, \mathcal{U}^\sigma_{TV}(\hat{P}^o)}}}\left[\sum_{s'}\left(P^{\pi, \mathcal{U}^\sigma_{TV}(P^o)} - P^{\pi, \mathcal{U}^\sigma_{TV}(\hat{P}^o)} \right)V^{\pi, \sigma}(s')\right] \\
    & \geq -\frac{\gamma}{1 - \gamma}\mathbb{E}_{d^\pi_{P^{\pi, \mathcal{U}^\sigma_{TV}(\hat{P}^o)}}}\left[\sum_{s'}\left|P^{\pi, \mathcal{U}^\sigma_{TV}(P^o)} - P^{\pi, \mathcal{U}^\sigma_{TV}(\hat{P}^o)} \right|\frac{r_{max}}{1-\gamma}\right] \\
    & \geq - \frac{2\gamma r_{max}}{(1-\gamma)^2}\mathbb{E}_{d^\pi_{P^{\pi, \mathcal{U}^\sigma_{TV}(\hat{P}^o)}}}\left[D_{TV}\left(P^{\pi, \mathcal{U}^\sigma_{TV}(P^o)},  P^{\pi, \mathcal{U}^\sigma_{TV}(\hat{P}^o)}\right)\right].
\end{aligned}
\end{equation}
For term $(b)$ in the RHS, based on the telescoping lemma \cite{luoalgorithmic}, we have:
\begin{equation}
\begin{aligned}
&\mathbb{E}_{s \sim d_0}[V^{\pi, \sigma}_{\hat{P}^o}(s)] - \mathbb{E}_{s \sim d_0}[V^{\pi, src}(s)] \\&= \frac{\gamma}{1 - \gamma}\mathbb{E}_{d^\pi_{P_{src}}}\left[\mathbb{E}_{P^{\pi, \mathcal{U}^\sigma_{TV}(\hat{P}^o)}}[V^{\pi, \sigma}_{\hat{P}^o}(s')] - \mathbb{E}_{P_{src}}[V^{\pi, \sigma}_{\hat{P}^o}(s')]\right] \\
&= \frac{\gamma}{1 - \gamma}\mathbb{E}_{d^\pi_{P_{src}}}\left[\inf_{P \in \mathcal{U}^{\sigma}(\hat{P}^o)}\mathbb{E}_{P}[V^{\pi, \sigma}_{\hat{P}^o}(s')] - \mathbb{E}_{P_{src}}[V^{\pi, \sigma}_{\hat{P}^o}(s')] \right] \\
&\geq -\frac{\gamma}{1 - \gamma}\mathbb{E}_{d^\pi_{P_{src}}}\left|\mathbb{E}_{P_{src}}[V^{\pi, \sigma}_{\hat{P}^o}(s')] - \inf_{P \in \mathcal{U}^{\sigma}(\hat{P}^o)}\mathbb{E}_{P}[V^{\pi, \sigma}_{\hat{P}^o}(s')]\right|.
\end{aligned}
\end{equation}
Thus, we have:
\begin{equation}
\begin{aligned}
   & \mathbb{E}_{s \sim d_0}[V^{\pi, \sigma}(s)] - \mathbb{E}_{s \sim d_0}[V^{\pi, src}(s)] \geq  \\&-\frac{2\gamma r_{max}}{(1-\gamma)^2}\mathbb{E}_{d^\pi_{P^{\pi, \mathcal{U}^\sigma_{TV}(\hat{P}^o)}}}\left[D_{TV}\left(P^{\pi, \mathcal{U}^\sigma_{TV}(P^o)},  P^{\pi, \mathcal{U}^\sigma_{TV}(\hat{P}^o)}\right)\right] \\
    &- \frac{\gamma}{1 - \gamma}\mathbb{E}_{d^{\pi}_{P_{src}}}\left[\left|\mathbb{E}_{P_{src}}[V^{\pi, \sigma}_{\hat{P}^o}(s')] - \inf_{P \in \mathcal{U}^{\sigma}(\hat{P}^o)}\mathbb{E}_{P}[V^{\pi, \sigma}_{\hat{P}^o}(s')]\right|\right].
\end{aligned}
\end{equation}
Or:
\begin{equation}
\begin{aligned}
&\mathbb{E}_{s \sim d_0}[V^{\pi, \sigma}(s)] \\&\geq \mathbb{E}_{s \sim d_0}[V^{\pi, src}(s)] -\frac{2\gamma r_{max}}{(1-\gamma)^2}\mathbb{E}_{d^\pi_{P^{\pi, \mathcal{U}^\sigma_{TV}(\hat{P}^o)}}}\left[D_{TV}\left(P^{\pi, \mathcal{U}^\sigma_{TV}(P^o)},  P^{\pi, \mathcal{U}^\sigma_{TV}(\hat{P}^o)}\right)\right] \\&- \frac{\gamma}{1 - \gamma}\mathbb{E}_{d^{\pi}_{P_{src}}}\left[\left|\mathbb{E}_{P_{src}}[V^{\pi, \sigma}_{\hat{P}^o}(s')] - \inf_{P \in \mathcal{U}^{\sigma}(\hat{P}^o)}\mathbb{E}_{P}[V^{\pi, \sigma}_{\hat{P}^o}(s')]\right|\right],
\end{aligned}
\end{equation}
which completes the proof.

We then present the convergence guarantee of the value Q function under the hybrid cross-domain robust RL setting. We first denote the normal Bellman operator for the \emph{source} domain is defined as $\mathcal{T}Q(s, a) = r(s, a) + \mathbb{E}_{s'\sim P_{src}}[\max_{a'}Q(s', a')], \forall s, a \in \mathcal{S}\times\mathcal{A}$, and the empirical of its as $\hat{\mathcal{T}}Q(s, a)$ that only backups single source transitions. We also denote the operator 
$\mathcal{T}^{\sigma, g}Q(s, a) = r(s, a) - \gamma\big(\mathbb{E}_{P^o}[(g(s, a) - \max_{a'}Q(s', a'))_{+}] - g(s, a)(1-\sigma)\big), \forall s, a \in \mathcal{S}\times\mathcal{A}$ and its empirical $\hat{\mathcal{T}}^{\sigma, g}$ as similar in \cite{panaganti2022robust}. We recall the offline nominal dataset $\mathcal{D}$ and source dataset $\mathcal{D}_{src}$. We consider the following approach using both source and target data:
\begin{equation}\label{eq:Q_update_general}
\begin{aligned}
 Q^{k+1} &\leftarrow \argmin_{Q} \kappa \mathbb{E}_{s, a, s' \sim \mathcal{D}}[(\hat{\mathcal{T}}^{\sigma, g} Q^k - Q)^2]+(1-\kappa) \mathbb{E}_{s, a, s' \sim \mathcal{D}_{src}}[(\hat{\mathcal{T}} Q^k - Q)^2],
\end{aligned}
\end{equation}
where $\kappa \in [0, 1]$ is the combination weight, $k$ denotes training iteration, and $\hat{\mathcal{T}}$ is the empirical Bellman operator. We denote $\mu$ and $\nu$ as the state-action distributions of target and source datasets. To maintain simplicity, we assume the source and the target datasets have the same state-action distribution, i.e. $\mu(s, a) = \nu(s, a), \forall s, a \in \mathcal{S}\times\mathcal{A}$. This assumption can hold easily when the source data $D_{src}$ is generated via a simulator, as it allows flexibility in selecting the transition starting points. We note that the source and target dynamics remain distinct ($P_{src} \neq P^o$).

We begin with the following proposition and lemma. 
\begin{prop}
    Denote $Q^{k+1}$ is the solution of Eq (\ref{eq:Q_update_general}) at the iteration $k+1$. Then at each iteration ($k=1, 2, \dots)$, we have:
    \begin{equation}
        Q^{k+1}
        (s, a)= \kappa \mathcal{T}^{\sigma, g}Q^k(s, a) + (1 - \kappa)\mathcal{T} Q^k(s, a). \forall s, a \in \mathcal{S} \times \mathcal{A}.
    \end{equation}
\end{prop}

\emph{Proof:} At the iteration $k+1$, we have $Q^{k+1}$ is the solution of the following:
\begin{equation}
\resizebox{1.0\textwidth}{!}{
    $
\begin{aligned}
 Q^{k+1} &\leftarrow \arg\min_{Q} \kappa \mathbb{E}_{\big(s, a, s^{\prime}\big) \sim \mathcal{D}}\big[\big(\hat{\mathcal{T}}^{\sigma, g} Q^{k} - Q\big)^2\big] + (1-\kappa) \mathbb{E}_{\big(s, a, s^{\prime}\big) \sim \mathcal{D}_{src}}\big[\big(\hat{\mathcal{T}} Q^k - Q\big)^2\big],
\end{aligned}
$}
\label{eq:q_update_proof}
\end{equation}
Setting the derivative of Eq (\ref{eq:q_update_proof}) w.r.t Q to zero at each $(s, a)$, we have:
\begin{equation}
    \begin{aligned}
        &\kappa \mathbb{E}_{P^o}\left[\mu(\hat{\mathcal{T}}^{\sigma, g}Q^k(s, a) - Q(s, a))\right] + (1 - \kappa)\mathbb{E}_{P_{src}}\left[\nu(\hat{\mathcal{T}}Q^k(s, a) - Q(s, a))\right] = 0.
    \end{aligned}
\end{equation}
Thus, we have:
\begin{equation}
    \begin{aligned}
        \kappa \mu Q(s, a) + (1 - \kappa) \nu Q(s, a) &= \kappa\mathbb{E}_{P^o}\left[\mu\hat{\mathcal{T}}^{\sigma, g}Q^k(s, a)\right] + (1-\kappa)\mathbb{E}_{P_{src}}\left[\nu\hat{T}Q^k(s, a)\right] \\
        &\stackrel{(a)}{=}\kappa\mu\mathcal{T}^{\sigma, g}Q^k(s, a) + (1 - \kappa)\nu\mathcal{T}Q^k(s, a),
    \end{aligned}
\end{equation}
where $(a)$ follows the fact that $\mathbb{E}_{P^o}[\hat{\mathcal{T}}^{\sigma, g}Q^k(s, a)] = \mathcal{T}^{\sigma, g}Q^k(s, a)$, and $\mathbb{E}_{P_{src}}$ $[\hat{\mathcal{T}}Q^k(s, a)] = \mathcal{T}Q^k(s, a)$.

We have $\mu = \nu$, thus we obtain value function at the next iteration $Q^k+1$ as follows:
\begin{equation}
    Q^{k+1}(s, a) = \kappa \mathcal{T}^{\sigma, g}Q^{k}(s,a) + (1 - \kappa)\mathcal{T}Q^k(s,a), \forall s, a \in \mathcal{S}\times\mathcal{A},
\end{equation}
which completes the proof.
\begin{lemma}
\label{lemma:lemma_q}
    Let define
    \begin{equation}
    \begin{aligned}
        &\xi = \max_{Q}\max_{s, a \in \mathcal{S}\times\mathcal{A}}\left|\mathcal{T}^{\sigma, g}Q(s, a) - \mathcal{T}^{\sigma}Q(s, a)\right|, \\ &\zeta = \max_Q\max_{s, a\in\mathcal{S}\times\mathcal{A}}\left|\mathcal{T}Q(s, a) - \mathcal{T}^{\sigma}Q(s, a)\right|.
    \end{aligned}
    \end{equation}
    For all iteration ($k=1, 2, \dots$), $Q^{k+1}$ is the solution of Eq (\ref{eq:Q_update_general}), we have the following holds:
    \begin{equation}
        \left\|Q^{k+1} - \mathcal{T}^{\sigma}Q^k \right\|_{\infty} \leq \kappa\xi + (1-\kappa)\zeta.
    \end{equation}
\end{lemma} 

\emph{Proof:} For any $s, a \in \mathcal{S}\times \mathcal{A}$, we have
\begin{equation}
    \begin{aligned}
        &Q^{k+1}(s, a) - \mathcal{T}^{\sigma}Q^k(s, a) \\&\leq \left| Q^{k+1}(s, a) - \mathcal{T}^{\sigma}Q^k(s, a) \right| \\
        & = \left| \kappa \mathcal{T}^{\sigma, g}Q^{k}(s, a) + (1 - \kappa)\mathcal{T}Q^k(s,a) - \mathcal{T}^{\sigma}Q^k(s, a) \right| \\
        &= \left| \kappa(\mathcal{T}^{\sigma, g}Q^{k}(s, a) - \mathcal{T}^{\sigma}Q^k(s, a)) + (1-\kappa)(\mathcal{T}Q^k(s,a) -\mathcal{T}^{\sigma}Q^k(s, a)) \right| \\
        & \stackrel{(a)}{\leq} \kappa\left| \mathcal{T}^{\sigma, g}Q^{k}(s, a) - \mathcal{T}^{\sigma}Q^k(s, a) \right| + (1-\kappa) \left| \mathcal{T}Q^k(s,a) -\mathcal{T}^{\sigma}Q^k(s, a) \right| \\
        &\leq \kappa \max_{Q}\max_{s, a \in \mathcal{S}\times\mathcal{A}}\left|\mathcal{T}^{\sigma, g}Q(s, a) - \mathcal{T}^{\sigma}Q(s, a)\right| \\&+ (1-\kappa) \max_Q\max_{s, a\in\mathcal{S}\times\mathcal{A}}\left|\mathcal{T}Q(s, a) - \mathcal{T}^{\sigma}Q(s, a)\right|,
    \end{aligned}
\end{equation}
where $(a)$ follows the triagle inequality. 
Thus, we have 
\begin{equation}
    \begin{aligned}
         &\left\|Q^{k+1} - \mathcal{T}^{\sigma}Q^k \right\|_{\infty} \\&\leq  \kappa \max_{Q}\max_{s, a \in \mathcal{S}\times\mathcal{A}}\left|\mathcal{T}^{\sigma, g}Q(s, a) - \mathcal{T}^{\sigma}Q(s, a)\right| \\&+ (1-\kappa) \max_Q\max_{s, a\in\mathcal{S}\times\mathcal{A}}\left|\mathcal{T}Q(s, a) - \mathcal{T}^{\sigma}Q(s, a)\right| \\
         & = \kappa\xi + (1-\kappa)\zeta,
    \end{aligned}
\end{equation}
which completes the proof.

Now, we provide the proof for Theorem \ref{theorem:convergence}.
\begin{restatable}[Convergence]{theorem}{convergence}
\label{theorem:convergence}
    Let $Q^*$ denote the optimal robust value function for the RMDP of the nominal model $P^o$, and define $Q^0 = 0$. Denote 
    \begin{equation}
    \begin{aligned}
        &\xi = \max_{Q}\max_{s, a \in \mathcal{S}\times\mathcal{A}}\left|\mathcal{T}^{\sigma, g}Q(s, a) - \mathcal{T}^{\sigma}Q(s, a)\right|,\\ 
        &\zeta = \max_{Q}\max_{s, a\in\mathcal{S}\times\mathcal{A}}\left|\mathcal{T}Q(s, a) - \mathcal{T}^{\sigma}Q(s, a)\right|.
    \end{aligned}
    \end{equation}
    Assume $\mu(s, a) = \nu(s, a), \forall s, a \in \mathcal{S}\times\mathcal{A}$, we have the following result holds:
    \begin{equation}
    \label{eq:q_converge_gap}
        \left\| Q^* - Q^{k+1} \right\|_{\infty}\leq \frac{\gamma^{k+1}r_{max}}{1 - \gamma} + \frac{1 - \gamma^{k+1}}{1-\gamma}\big(\kappa\xi + (1-\kappa)\zeta\big).
    \end{equation}
\end{restatable}

\emph{Proof:} We have 
\begin{equation}
    \begin{aligned}
        &\left\| Q^* - Q^{k+1} \right\|_{\infty}
        \\& = \left\| Q^* - \mathcal{T}^{\sigma}Q^k + \mathcal{T}^{\sigma}Q^k - Q^{k+1}\right\|_{\infty} \\
        &\stackrel{(a)}{\leq} \left\| Q^* - \mathcal{T}^{\sigma}Q^k  \right\|_{\infty} + \left\| Q^{k+1} - \mathcal{T}^{\sigma}Q^k \right\|_{\infty} \\
        &\stackrel{(b)}{\leq} \gamma \left\| Q^* - Q^k \right\|_{\infty} + \left\| Q^{k+1} - \mathcal{T}^{\sigma}Q^k \right\|_{\infty} \\
        &\stackrel{(c)}{\leq} \gamma^{k+1} \left\| Q^* - Q^0 \right\|_{\infty} + \gamma^k \left\| Q^1 - \mathcal{T}^{\sigma}Q^0 \right\|_{\infty} \\& + \dots + \gamma \left\| Q^k - \mathcal{T}^{\sigma}Q^{k-1} \right\|_{\infty} + \left\| Q^{k+1} - \mathcal{T}^{\sigma}Q^k \right\|_{\infty},
    \end{aligned}
\end{equation}
where $(a)$ follows the triangle inequality, $(b)$ follows that $Q^*$ is the fixed point of the robust Bellman operator, and the robust Bellman operator is a contraction mapping w.r.t the infinity norm, and $(c)$ is the results when we recursively apply the previous inequalities.

Using Lemma \ref{lemma:lemma_q}, we obtain the following:
\begin{equation}
    \begin{aligned}
       \left\| Q^* - Q^{k+1} \right\|_{\infty} &\leq  \gamma^{k+1} \left\| Q^* - Q^0 \right\|_{\infty} + \gamma^k \left( \kappa\xi + (1-\kappa)\zeta \right) \\&+ \dots + \gamma \left( \kappa\xi + (1-\kappa)\zeta \right) + \left( \kappa\xi + (1-\kappa)\zeta \right) \\ 
       &\leq \frac{\gamma^{k+1}r_{max}}{1-\gamma} + \frac{1 - \gamma^{k+1}}{1 - \gamma} \left( \kappa\xi + (1-\kappa)\zeta \right),
    \end{aligned}
\end{equation}
where the last inequality follows the fact that $Q^0 = 0$ and $Q^*\leq \frac{r_{max}}{1-\gamma}$.

\section{Algorithm Description}\label{sec:alg_des}
\begin{algorithm}[t]
\caption{HYbrid cross-Domain RObust RL - HYDRO}
\label{alg:robust}
\begin{algorithmic}[1]
\label{alg:hydro}
\STATE\textbf{Input:} Offline dataset $\mathcal{D}$, radius of robustness $\sigma$, maximum perturbation $\Phi$, target update rate $\tau$, mini-batch size $N$, number of actions $u$,  ensemble size $E$, source $\mathcal{M}_{src}$, source replay buffer $\mathcal{D}_{src}$, uncertainty  coefficient $\alpha$.
\STATE \textbf{Initialize:} Two Q networks $Q_{\phi_1}$ and $Q_{\phi_2}$, one dual neural network $g_\theta$, policy (perturbation) model: $\pi_{\varepsilon} \in [-\Phi, \Phi]$, and action VAE $G^{a}_\Theta$, and target networks $Q_{\phi'_1}, Q_{\phi'_2}, \pi_{\varepsilon'}$.
\STATE Train the ensemble transition model $\hat{P}^o = \{\hat{P}^o_i(s'|s, a) = \mathcal{N}\left(\mu_{\varphi}(s, a), \Sigma_{\varphi}(s, a)\right)\}^E_{i=1}\}$ on $\mathcal{D}$:
\STATE Compute uncertainty threshold $\epsilon_u = \frac{1}{\alpha}\max_{s, a \in \mathcal{D}}u(s, a)$.

\FOR {$k = 1, \dots, \text{K}$} 
    \STATE Rollout $h$ steps with $\mathcal{M}_{src}$.
    \FOR{$i=1, \dots, h$} 
        \STATE Compute the uncertainty for each source transition $u_i(s_i, a_i) = \max_{j, k}\|\mu^j_{\varphi}(s_i, a_i) - \mu^k_{\varphi}(s_i, a_i)\|^2$.
        \IF{$u_i \leq \epsilon_u$} 
            \STATE $\mathcal{D}_{src} \leftarrow \mathcal{D}_{src} \bigcup (s_i, a_i, r_i, s'_i)$. 
        \ENDIF
    \ENDFOR
    \STATE Sample a batch target data $B_{tar} = \{(s, a, r, s')^i_{tar}\}^{N}_{i=1}$ uniformly from $\mathcal{D}$.
    \STATE Sample a batch source data $B_{src} = \{(s, a, r, s')^i_{src}\}^{N}_{i=1}$ with probability $p^i(s, a, s')$ from $D_{src}$.
    \STATE Denote $B = B_{tar} \cup B_{src}.$
    \STATE Train the VAE $w \leftarrow \arg \min _{\omega}ELBO(B;G^{a}_\Theta)$. Sample $u$ action $a'_i$ from $G^{a}_\Theta(s')$ for each $s' \in B$.
    \STATE Perturb $u$ actions $a'_i = a'_i + \pi_{\varepsilon}(s', a'_i)$ for each $s' \in B$.
    \STATE Compute $V(s')$ for each $s'$ in $B$:
    \begin{equation}
        V(s') = \max_{a'_i}(0.75 \min\{Q_{\theta'_1}, Q_{\theta'_2}\} + 0.25\max\{Q_{\theta'_1}, Q_{\theta'_2}\}). 
    \end{equation}
    \STATE Compute $Q_{tar}(s, a)$ for each $(s, a, r, s')$ in $B_{tar}$:
    \begin{equation}
        Q_{tar}(s, a) = r - \gamma \max\{g_\theta(s, a) - V(s'), 0\} + \gamma(1-\sigma)g_\theta(s, a).
    \end{equation}
    \STATE Compute $Q_{src}(s, a)$ for each $(s, a, s', r)$ in $B_{src}$:
    $Q_{src}(s, a) = r(s, a) + \gamma V(s')$. 
    \STATE Compute $Q(s, a)$ for each $(s_{src}, a_{src}) \in \mathcal{D}_{src}, \hat{s}'_{i} \sim \hat{P}^o(.|s_{src}, a_{src})$.
    \STATE Compute and update transition priority $\psi(s, a)$ for each transition in $B_{src}$.
    \STATE Update dual network $g_\theta$ using $(s_{tar}, a_{tar}, s'_{tar})$ and $(s_{src}, a_{src}, s'_{tar})$, where ${s'_{tar} \sim \hat{P}^o(.|s_{src}, a_{src})}$:
    \begin{equation}
        \theta_3 \leftarrow \arg \min_{\theta}\sum[\max\{g_\theta(s, a) -  V_{t}(s'), 0\} - (1-\sigma)g_\theta(s, a)].
    \end{equation}
    \STATE Update Q function using both target and simulation data 
    \begin{equation}
    \begin{aligned}
        Q_{\phi} \leftarrow &\arg \min_{Q_\phi}\big(\sum^{B_{tar}}\left(Q_{\text{tar}}(s, a) - Q_\phi(s, a)\right)^2 \\&+ \sum^{B_{src}}( \mathds{1}(\psi(s, a) > \psi_{k\%}))\left(Q_{\text{sim}}(s, a) - Q_\phi(s, a)\right)^2\big).
    \end{aligned}
    \end{equation}
    
    \STATE Update policy to maximize the Q network using the data from both source and target batches.
    
    \STATE Update target network: $\theta' = (1-\tau)\theta' + \tau\theta, \varepsilon' = (1-\tau)\varepsilon'+\tau\varepsilon$.
\ENDFOR
\STATE \textbf{return} $Q_\phi$.
\end{algorithmic}
\end{algorithm}
We present the pseudocode of HYDRO in Algorithm \ref{alg:hydro}.
\section{Experimental Details and Hyperparameters}\label{sec:exp_details}
This section provides the implementation details and hyperparameters. To facilitate reproducibility, we have included HYDRO's code in the supplementary material and will open-source our code upon acceptance.
\subsection{Environment Setting}\label{sec:appendix_env_settings}
\paragraph{Dataset}
We conduct our experiments on three MuJoCo environments (HalfCheetah, Walker2d, Hopper). For the offline target dataset, we use the Medium version 2 datasets from D4RL \cite{fu2020d4rl} as our offline datasets \cite{panaganti2022robust}. This aligns with the setting for the target datasets in H2O \cite{niu2022trust}, which also considered the offline target online source in the standard RL setting. 
To create the scarce data settings, we use 10\% of these datasets for training, i.e. 100K target transitions from the D4RL dataset.
\begin{figure}[h]
    \centering
    \includegraphics[width=1.0\linewidth]{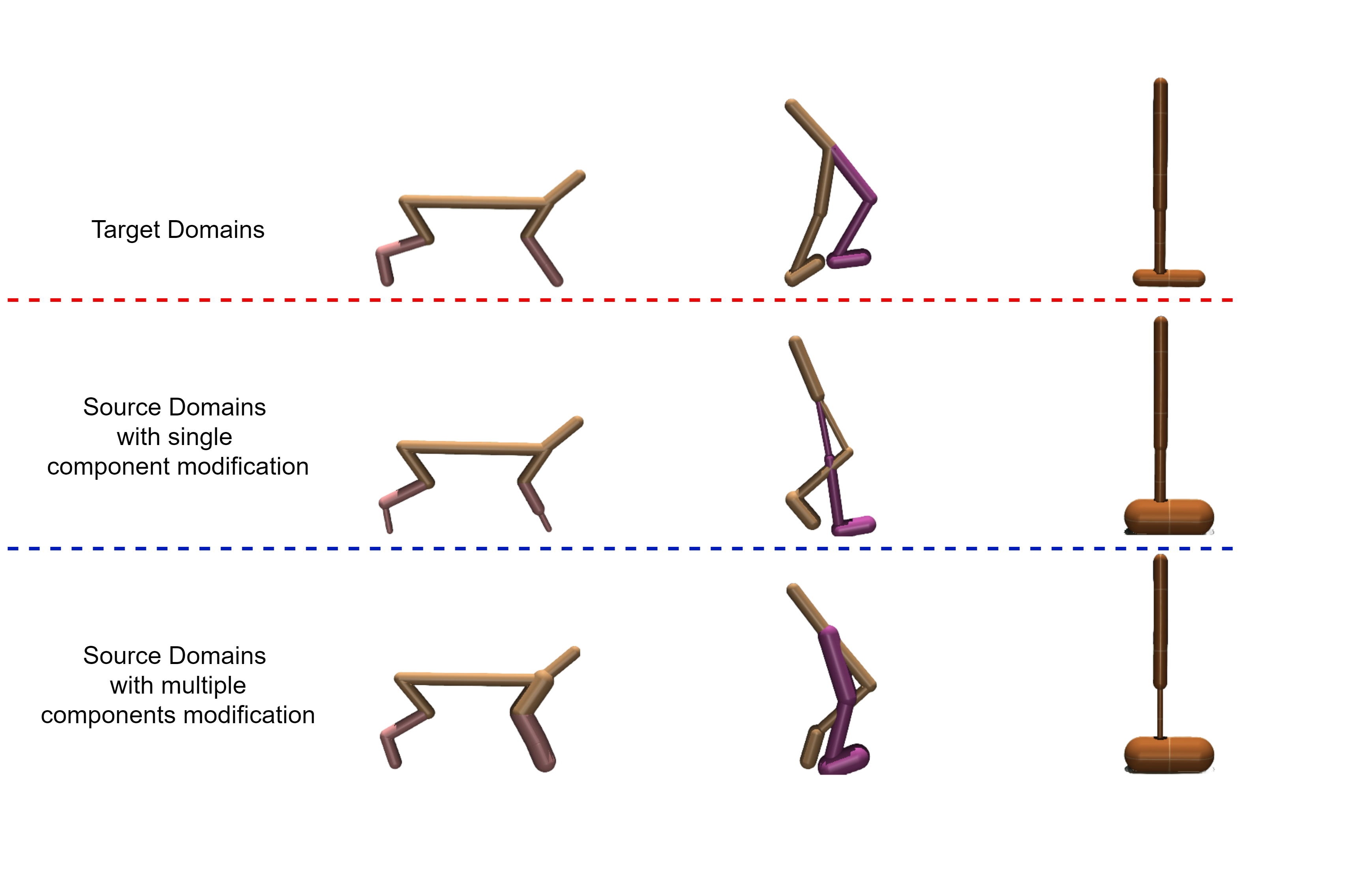}
    \caption{Graphical illustration of all environments. The target domains (top) are well-functional simulated robots from Mujoco Gym, and the source domains have morphology shifts via single (middle) or multiple (bottom) component modifications.}
    \label{fig:envs}
\end{figure}
\paragraph{Source Environment} 
While H2O modifies source environment parameters like gravity and friction, we modify the agent morphology to realise our robust setting benchmark. This is done to prevent any information leakage from the source data. 
For each Mujoco benchmark, we propose two variants with morphology shifts as our source environments. Specifically, we consider two types of modifications: single-comp, in which we modify a single type of component of the agent, and multi-comp, in which we alter multiple components of the agent. This yields a total of six source domains, with environment illustrations provided in Figure \ref{fig:envs}. We use 1 million source data for all methods. Finally, we provide the detailed modifications of the environments as the following:

\textbf{HalfCheetah - single comp}: We modify the size of the feet. Detailed modifications of the XML file are:
\begin{lstlisting}[language=python]
<geom axisangle="0 1 0 -.27" name="bfoot" pos=".03 0 -.097" rgba="0.9 0.6 0.6 1" size="0.023 .094" type="capsule"/>
\end{lstlisting}
\begin{lstlisting}[language=python]
<geom axisangle="0 1 0 -.6" name="ffoot" pos=".045 0 -.07" rgba="0.9 0.6 0.6 1" size="0.023 .07" type="capsule"/>
\end{lstlisting}

\textbf{HalfCheetah - multi comp}: We modify the size of the front components (thigh, leg, foot). Detailed modifications of the XML file are:
\begin{lstlisting}[language=python]
<geom axisangle="0 1 0 .52" name="fthigh" pos="-.07 0 -.12" size="0.08 .133" type="capsule"/>

\end{lstlisting}
\begin{lstlisting}[language=python]
  <geom axisangle="0 1 0 -.6" name="fshin" pos=".065 0 -.09" rgba="0.9 0.6 0.6 1" size="0.08 .106" type="capsule"/>
\end{lstlisting}
\begin{lstlisting}[language=python]
<geom axisangle="0 1 0 -.6" name="ffoot" pos=".045 0 -.07" rgba="0.9 0.6 0.6 1" size="0.08 .07" type="capsule"/>
\end{lstlisting}

\textbf{Walker2d - single comp}: We modify the size of the thighs. Detailed modifications of the XML file are:
\begin{lstlisting}[language=python]
<geom friction="0.9" fromto="0 0 1.05 0 0 0.6" name="thigh_geom" size="0.025" type="capsule"/>
\end{lstlisting}
\begin{lstlisting}[language=python]
<geom friction="0.9" fromto="0 0 1.05 0 0 0.6" name="thigh_left_geom" rgba=".7 .3 .6 1" size="0.025" type="capsule"/>
\end{lstlisting}

\textbf{Walker2d - multi comp}: We modify the size of components in one side (thigh, leg, foot). Detailed modifications of the XML file are:
\begin{lstlisting}[language=python]
<geom friction="0.9" fromto="0 0 1.05 0 0 0.6" name="thigh_left_geom" rgba=".7 .3 .6 1" size="0.075" type="capsule"/>
\end{lstlisting}
\begin{lstlisting}[language=python]
  <geom friction="0.9" fromto="0 0 0.6 0 0 0.1" name="leg_left_geom" rgba=".7 .3 .6 1" size="0.06" type="capsule"/>
\end{lstlisting}
\begin{lstlisting}[language=python]
<geom friction="1.9" fromto="-0.0 0 0.1 0.2 0 0.1" name="foot_left_geom" rgba=".7 .3 .6 1" size="0.09" type="capsule"/>
\end{lstlisting}

\textbf{Hopper - single comp}: We modify the size of the foot. Detailed modifications of the XML file are:
\begin{lstlisting}[language=python]
<geom friction="2.0" fromto="-0.13 0 0.1 0.26 0 0.1" name="foot_geom" size="0.12" type="capsule"/>
\end{lstlisting}

\textbf{Hopper - multi comp}: We modify the size of the leg and foot. Detailed modifications of the XML file are:
\begin{lstlisting}[language=python]
<geom friction="0.9" fromto="0 0 0.6 0 0 0.1" name="leg_geom" size="0.02" type="capsule"/>
\end{lstlisting}
\begin{lstlisting}[language=python]
 <geom friction="2.0" fromto="-0.13 0 0.1 0.26 0 0.1" name="foot_geom" size="0.12" type="capsule"/>
\end{lstlisting}

\subsection{Algorithm’s Implementations}\label{sec:appendix_algo_imp}
This section provides the implementation details and hyperparameters of all methods. 


\paragraph{RFQI} We follow the implementation of the RFQI algorithm from the official code (\url{https://github.com/zaiyan-x/RFQI}). Following their approach, we do a grid search for the radius of the uncertainty set $\sigma \in \{0.2, 0.3, \dots, 0.8\}$ and the learning rate and report their best performances. We find that $\sigma=0.3$ is the best for HalfCheetah, and $\sigma=0.7$ is the best for Walker2d and Hopper. The best learning rates are reported as the following: $[1e^{-3}, 1e^{-3}]$ for HalfCheetah and Walker2d, and $[3e^{-4}, 3e^{-4}]$ for Hopper. Finally, we run RFQI with the default configuration reported in \cite{panaganti2022robust}. Furthermore, we trained \emph{RFQI} with a full offline target dataset, which we refer to as \emph{Oracle}.

\paragraph{HYDRO} In our experiments, we use an ensemble of $N$ dynamics model $\{\hat{P}^o_i(s'|s,$ $ a) = \mathcal{N}\left(\mu_{\varphi}(s, a), \Sigma_{\varphi}(s, a)\right)\}^N_{i=1}$. We use 5 MDP layers with 200 units each layer. Each layer in the network uses Swish activation following prior works \cite{chua2018deep,yu2020mopo}. We set the ensemble size to 7. Each model in the ensemble is trained using offline target dataset $\mathcal{D}_{tar}$ via the maximum log-likelihood:
\begin{equation}
    \mathcal{L}_{\varphi}=\mathbb{E}_{\left(s, a, r, s'\right) \sim \mathcal{D}}\left[-\log \hat{P}^o\left(s' \mid s, a\right)\right].
\end{equation}
During the training, we perform 2000 interactions with the source environment after every 1000 training iterations.
The uncertainty coefficient $\alpha$ determines the uncertainty threshold $\epsilon_u = \frac{1}{\alpha}\max_{s, a \in \mathcal{D}}u(s, a)$. Thus, it controls the uncertainty of source samples we add to the replay buffer for training policy. Specifically, we set $\alpha = \beta \times h$, where $h$ is the rollout length, and we use $h=5$ as following prior model-based offline RL method \cite{sun2023model,zhang2023uncertainty}. We observe that setting $\beta=2.0$ yields satisfactory performance on HalfCheetah and Walker2d environments. However, for the Hopper benchmark, which required more pessimistic samples \cite{zhang2023uncertainty}, we set $\beta=3.0$. The implementation of priority sampling is based on \cite{schaul2015prioritized} implemented in \cite{labml}.
We employ RFQI as our offline robust RL backbone with a discount factor $\gamma$ of 0.99 across all environments, a target smoothing coefficient $\tau$ of 0.005, and a batch size of 100. As we use RFQI as our backbone, we fix the hyperparameters related to robustness and learning rate as RFQI except for HalfCheetah - single comp, where we find setting $[3e^{-4}, 8e^{-4}]$ is slightly better, as shown in Table \ref{tab:lr_comparison}. Our method only introduces two hyperparameters, which are the top $k$ score selection and uncertainty filter $\beta$. Both the value function and policy are updated at each training step. 


\paragraph{H2O} We utilize the author's official implementation from their GitHub repository (\url{https://github.com/t6-thu/H2O}). We integrate our environments into the code and execute it using the default hyperparameters mentioned in the paper \cite{niu2022trust}.

\paragraph{PQL} We train PQL \cite{liu2020provably} (with filtration thresholding $b=0$), following the default configuration as outlined in \cite{liu2020provably,panaganti2022robust}.

Following \cite{panaganti2022robust}, we train all methods with the maximum interaction is $5\times 10^5$.

\begin{table}[t]
\centering
\caption{Robust performance comparison of HYDRO with two different learning rates in HalfCheetah - single comp.}
\label{tab:lr_comparison}
\begin{tabular}{@{}ccc@{}}
\toprule
                     & \multicolumn{2}{c}{Halfcheetah - Single comp}                 \\ \midrule
                     & Front joint stiffness & Back actuator ctrlrange \\
$[1e^{-3}, 1e^{-3}]$ & 4780 ± 569            & 4334 ± 300              \\
$[3e^{-4}, 8e^{-4}]$ & 4781 ± 476   & 4399 ± 382     \\ \bottomrule
\end{tabular}
\end{table}

\begin{table}[t]
\centering
\caption{Computing infrastructure.}
\label{tab:server}
\begin{tabular}{@{}ccccc@{}}
\toprule
\textbf{CPU}                   & \textbf{Number of CPU Cores} & \textbf{GPU} & \textbf{VRAM} & \textbf{RAM} \\ \midrule
Intel(R) Xeon(R) Gold 6248 CPU & 20                           & V100         & 32 GB         & 377 GB       \\ \bottomrule
\end{tabular}
\end{table}

\subsection{Computing Infrastructure}
We use Python 3.7.12, Pytorch 1.13.0, Gym 0.23.1, Mujoco 2.3.2, and D4RL 1.1. All experiments are conducted on a Ubuntu 22.4 server with CUDA version 12.2. Finally, we provide the computing infrastructure that we use to run our experiments in Table \ref{tab:server}.

\section{Additional Experimental Results}\label{sec:add_exps}
\subsection{Statistical Significance}
To assess the statistical significance of our proposed method's performance compared to baselines, we conduct a z-test (given our sample size of 30 for each method) on the average returns reported in Table 1 of the main paper. Specifically, we perform multiple hypothesis testing, comparing each baseline to HYDRO. The tests aim to determine whether the average returns of the baseline ($R_{\text{baseline}}$) is significantly lower than our method HYDRO ($R_{\text{HYDRO}}$). The alternative hypothesis $H_a: R_{\text{baseline}} >  R_{\text{HYDRO}}$ is tested against the null hypothesis $H_0 : R_{\text{baseline}} \leq  R_{\text{HYDRO}}$. We provide the detailed results in Table \ref{tab:significant_test}. Notably, we observe that the p-values are significantly lower than 0.05 in 10 out of 12 tasks for both RFQI and PQL, suggesting that our method's improvement is statistically significant.

\begin{table}[t]
\centering
\caption{We conduct a Z-test to assess whether the average returns of the baselines are lower than our method HYDRO. The null hypothesis $H_0: R_{\text{baseline}} \leq  R_{\text{HYDRO}}$ and the alternative hypothesis $H_a: R_{\text{baseline}} >  R_{\text{HYDRO}}$. The p-value corresponding to each test is provided in each cell. ``-m": multi comp, ``-s": single comp.}
\label{tab:significant_test}
\begin{tabular}{@{}ccccccc@{}}
\toprule
\multicolumn{1}{l}{}         & \multicolumn{1}{l}{}                                                                        & \multicolumn{1}{l}{\textbf{}} & \textbf{RFQI} & \textbf{PQL} & \textbf{H2O-m} & \textbf{H2O-s} \\ \midrule
\multirow{4}{*}{HalfCheetah} & \multirow{2}{*}{\textit{\begin{tabular}[c]{@{}c@{}}Front joint\\ stiffness\end{tabular}}}   & HYDRO-m            & 3.44e-09      & 1.85e-15     & 8.85e-160                 & 7.61e-247                  \\
                             &                                                                                             & HYDRO-s           & 8.03e-32      & 2.97e-32     & 0.0                       & 0.0                        \\
                             & \multirow{2}{*}{\textit{\begin{tabular}[c]{@{}c@{}}Back actuator\\ ctrlrange\end{tabular}}} & HYDRO-m            & 5.38e-05      & 2.21e-40     & 2.93e-185                 & 2.17e-185                  \\
                             &                                                                                             & HYDRO-s           & 0.0022        & 7.43e-36     & 3.20e-183                 & 2.63e-180                  \\
\multirow{4}{*}{Walker2d}    & \multirow{2}{*}{\textit{Gravity}}                                                           & HYDRO-m            & 0.074         & 5.37e-11     & 1.40e-297                 & 0.0                        \\
                             &                                                                                             & HYDRO-s           & 3.52e-05      & 6.04e-26     & 0.0                       & 0.0                        \\
                             & \multirow{2}{*}{\textit{\begin{tabular}[c]{@{}c@{}}Foot joint\\ stiffness\end{tabular}}}    & HYDRO-m            & 1.73e-19      & 0.96         & 9.88e-122                 & 2.85e-261                  \\
                             &                                                                                             & HYDRO-s           & 4.80e-33      & 0.00482      & 9.82e-159                 & 0.0                        \\
\multirow{4}{*}{Hopper}      & \multirow{2}{*}{\textit{Gravity}}                                                           & HYDRO-m            & 3.40e-16      & 2.85e-149    & 0.0                       & 0.0                        \\
                             &                                                                                             & HYDRO-s           & 0.0222        & 6.18e-108    & 0.0                       & 0.0                        \\
                             & \multirow{2}{*}{\textit{\begin{tabular}[c]{@{}c@{}}Foot joint\\ stiffness\end{tabular}}}    & HYDRO-m            & 0.00076       & 0.00021      & 1.72e-235                 & 3.90e-98                   \\
                             &                                                                                             & HYDRO-s           & 0.51          & 0.59         & 3.27e-125                 & 2.33e-46                   \\ \cmidrule(l){1-7} 
\end{tabular}
\end{table}

\begin{figure}[H]

    \centering
    \includegraphics[width=1.0\linewidth]{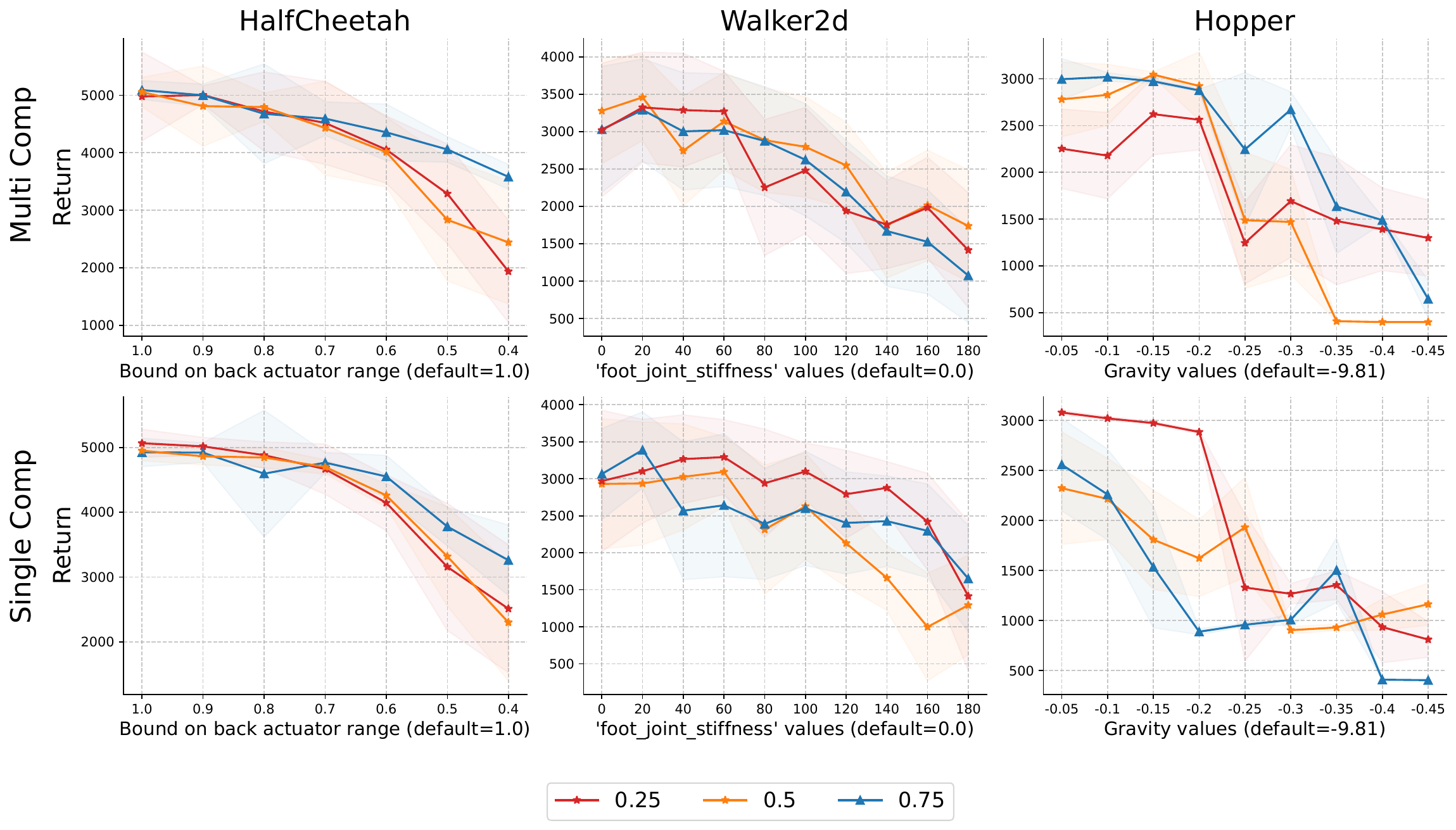}
    \caption{Parameter study for top $k$ score data selection. The lines represent the average returns over 30 different seeded runs, and the shaded areas represent the standard deviation.}
    \label{fig:topk}
\end{figure}
\subsection{Top $k$ Score Data Selection}

The optimal top $k$ score for data selection is influenced by the dynamics gap between source and target domains, varying across different task environments. We employ different top $k$ score data selections (25\%, 50\%, 75\%) for our algorithm. The results are shown in Figure \ref{fig:topk}. We observe that the HalfCheetah environment benefits from larger top $k$ selections (k=75\%), while Walker and Hopper tasks perform better with smaller selections (k = 50\%, 25\%). This emphasizes the need to tailor the top $k$ score data selection to achieve good performance.

\subsection{Uncertainty Filtering Data Ratio}\label{sec:appendix_unc_filter}
Intuitively, the uncertainty filtering data $\beta$ controls how reliable source data is for the target domain, with high value imposing more conservative. We conduct the experiments to examine the influence of the value of $\beta$ to HYDRO performance and report the results in Table \ref{tab:beta}. We observed that for Halfcheetah (HC) and Walker (WK), $\beta=2.0$ yields satisfactory performance. However, Hopper (HP) requires more pessimistic samples, and $\beta=3.0$ has the best performance. It is similar with the observation in previous offline RL work \cite{zhang2023uncertainty}

\begin{table}[ht]
\centering
\caption{Uncertainty filter $\beta$.}
\begin{tabular}{@{}cccc@{}}
\toprule
  & HC\_front\_joint\_stiff & WK\_gravity & HP\_gravity \\ \midrule
1.0 & 4317±1194               & 3335±238    & 1231±255    \\
2.0 & 4456±872                & 3225±487    & 1423±431    \\
3.0 & 4204±1122               & 3187±522    & 1858±468    \\ \bottomrule
\end{tabular}

\label{tab:beta}
\end{table}

\begin{figure}[t]
    \centering
    \includegraphics[width=0.5\linewidth]{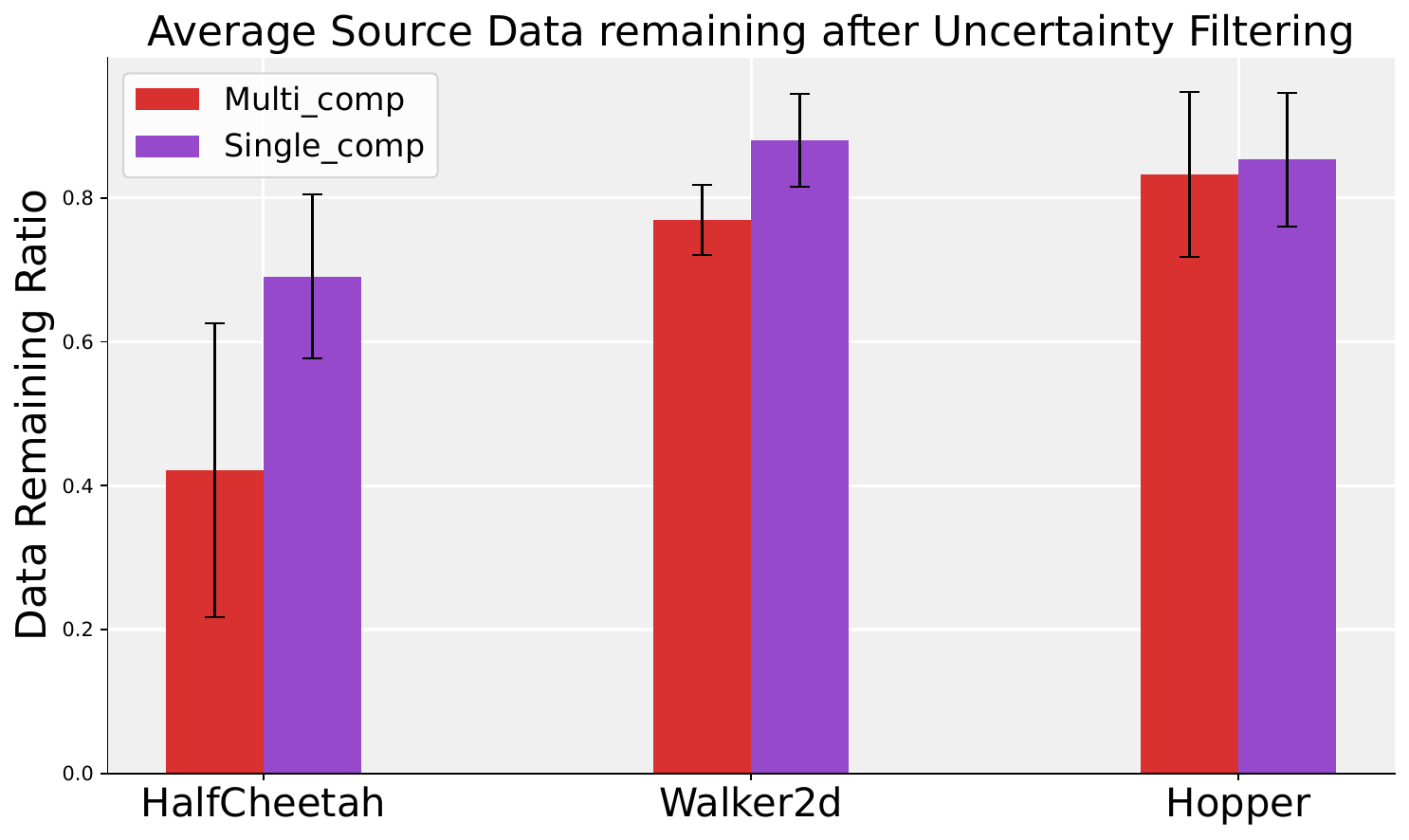}
    \caption{Average source data remaining ratio after performing uncertainty filtering.}
    \label{fig:unc_data_ratio}
\end{figure}

We further examine the effect of the uncertainty value on the number of source samples used for training. Figure \ref{fig:unc_data_ratio} visualizes the average source data remaining after uncertainty filtering. We observe a distinct gap in the remaining data ratio between single-component and multiple-component modifications. This suggests that source domains with multiple component modifications exhibit more `uncertain' transitions relative to the target domain compared to those with single-component modifications.

\subsection{HYDRO with different domain-gaps measurement.}
To emphasize the distinction between HYDRO and \emph{standard} cross-domain RL methods, we evaluate HYDRO's performance using an alternative domain-gaps measurement.
Specifically, we replace our measurement, which quantifies the discrepancy between the \emph{worst-case} target model and the source model, with the standard measurement from cross-domain RL, which quantifies the discrepancy between the target model and the source model. To implement this, we adopt the domain classifiers used in DARC \cite{eysenbachoff}. 
Table \ref{tab:hydro_with_different_measurement} illustrates this substitution leads to significant performance degradation. These results underscore the inadequacy of standard domain-gap measurement in \emph{robust} RL settings and highlight the critical role and effectiveness of HYDRO's tailored measurement approach.
\begin{table}[t]
\centering
\caption{Robust performance comparison between HYDRO and its variant using domain classifiers as the domain-gap metric. ``Target - Source" refers to measuring the gap between the target and source, while ``Worst-case target - Source" represents the gap between the worst-case target and source.}
\label{tab:hydro_with_different_measurement}
\begin{tabular}{@{}ccc@{}}
\toprule
                                                                                 & Target - Source & Worst-case target - Source \\ \midrule
\begin{tabular}[c]{@{}c@{}}Halfcheetah - Back actuator ctrlrange\end{tabular} & 3705 ± 735      & \textbf{4480 ± 350}        \\ \midrule
\begin{tabular}[c]{@{}c@{}}Walker2d - Gravity\end{tabular}                       & 3097 ± 473      & \textbf{3225 ± 487}        \\ \midrule
\begin{tabular}[c]{@{}c@{}}Hopper - Gravity\end{tabular}                         & 980 ± 392       & \textbf{1858 ± 468}        \\ \bottomrule
\end{tabular}
\end{table}

\subsection{How HYDRO performs under harder limited target data settings?}\label{sec:discussion_harder}
\begin{figure}[t]
    \centering
    \includegraphics[width=0.5\linewidth]{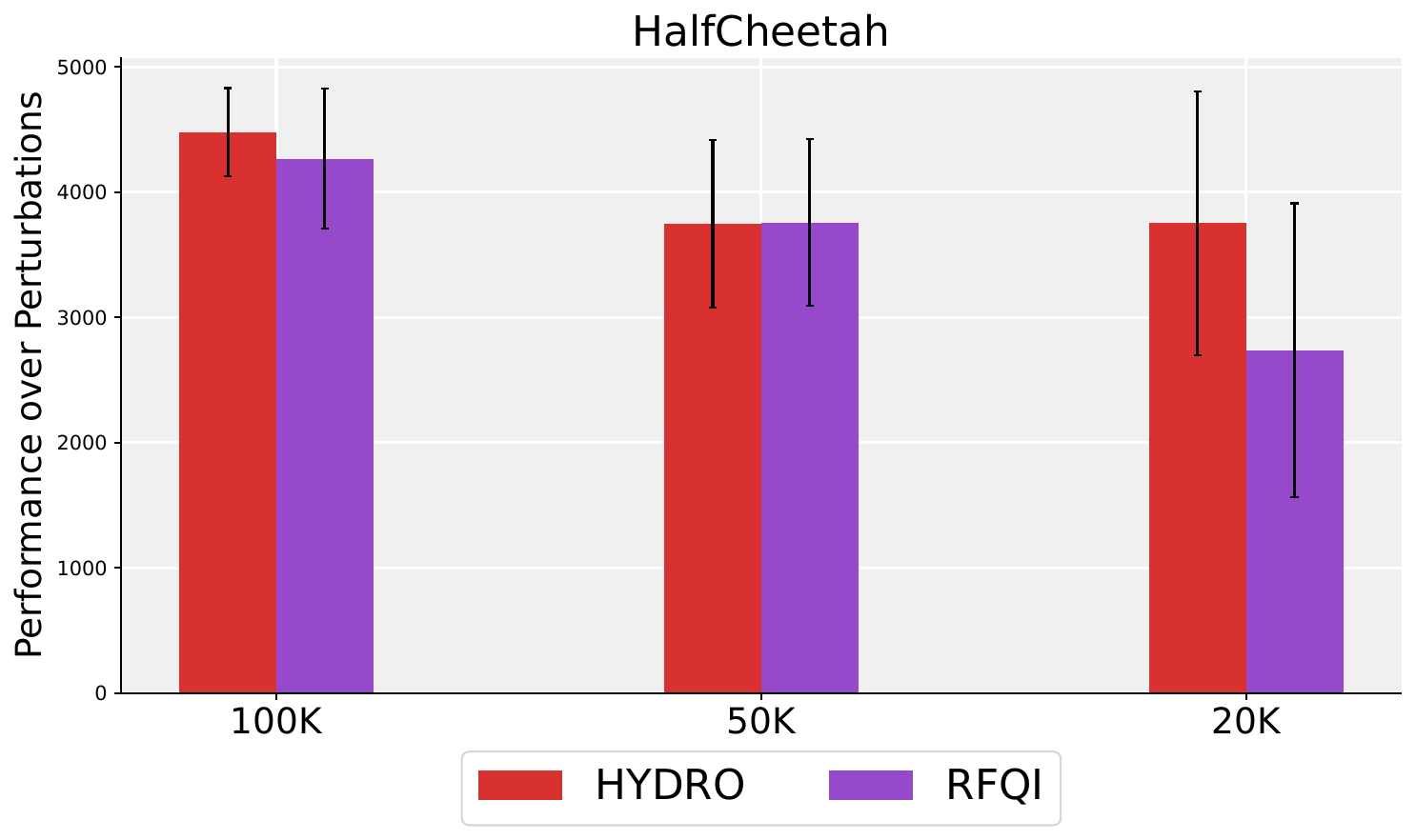}
    \caption{Robust performance comparison between HYDRO and RFQI under various limited target data settings.}
    \label{fig:HYDRO_various}
\end{figure}

To further understand H2O's performance, we analyze HYDRO's behavior under increasingly challenging, data-limited target settings. 
Figure \ref{fig:HYDRO_various} illustrates RFQI's robust performance decreases substantially as target data decreases, while HYDRO maintains consistently strong performance with only minimal degradation. These results demonstrate HYDRO's effectiveness in overcoming data scarcity challenges.

\subsection{Training time}\label{sec:appendix_training_time}
The training time for H2O, RFQI, and HYDRO are 14hr, 1 day 10hr and 1 day 12hr respectively. The training time for robust methods is longer due to the inner optimization. \textbf{Our method only needs slightly higher computational cost compared to RFQI, due to the training time of the transition dynamics model and data selection components.}

\subsection{Additional Experiments}\label{sec:appendix_add_exps}
In this section, we further compare the performance of HYDRO with the current state-of-the-art methods for cross-domain RL in the online source and online target setting, and single-domain method CQL \cite{kumar2020conservative}.  Table \ref{tab:par_vgdf_cql} illustrates that HYDRO performs the best compared to these baselines. We note that while our method focuses on \emph{offline} target setting, PAR \cite{lyucross} and VGDF \cite{xu2024cross} were designed for \emph{online} target setting thus affecting their performances.

\begin{table}[H]
\centering
\caption{Average returns over different environment parameter perturbations over 30 different seeded runs.}
\label{tab:par_vgdf_cql}
\begin{tabular}{@{}ccccc@{}}
\toprule
               & CQL                                                     & PAR                                                     & VGDF                                                    & HYDRO                                                            \\ \midrule
Hopper gravity &  903±117  & 664±217  & 910±194  & \textbf{1858±468} \\
Walker gravity &  2711±365 & 1727±781 &  2590±659 &  \textbf{3225±487} \\ \bottomrule
\end{tabular}
\end{table}

\end{document}